\documentclass[letterpaper, 10 pt, journal, twoside]{IEEEtran}  % Comment this line out if you need a4paper

\usepackage{float, caption}
\usepackage{amsmath, amssymb , graphicx, subcaption}
\usepackage{multirow}
\usepackage{dblfloatfix} 
\usepackage{balance}
\usepackage[ruled,vlined,linesnumbered]{algorithm2e}
\usepackage{xcolor}

\makeatletter

\DeclareRobustCommand\bigop[1]{%
  \mathop{\vphantom{\sum}\mathpalette\bigop@{#1}}\slimits@
}
\newcommand{\bigop@}[2]{%
  \vcenter{%
    \sbox\z@{$#1\sum$}%
    \hbox{\resizebox{\ifx#1\displaystyle.9\fi\dimexpr\ht\z@+\dp\z@}{!}{$\m@th#2$}}%
  }%
}

\newcommand{\bigVee}{\DOTSB\bigop{\vee}}
\makeatother

\title{\LARGE \bf 
Autonomous Robotic Suction to Clear the Surgical Field for Hemostasis using Image-based Blood Flow Detection}

\author{Florian Richter$^1$ \IEEEmembership{Student Member, IEEE}, Shihao Shen$^1$ \IEEEmembership{Student Member, IEEE}, Fei Liu$^1$, Jingbin Huang$^1$, \\Emily K. Funk$^2$, Ryan K. Orosco$^2$ \IEEEmembership{Member, IEEE}, and Michael C. Yip$^1$ \IEEEmembership{Member, IEEE}
\thanks{$^1$Florian Richter, Shihao Shen, Fei Liu, Jingbin Huang, and Michael C. Yip are with the Department of Electrical and Computer Engineering, University of California San Diego, La Jolla, CA 92093 USA. {\tt\small \{frichter, s1shen, f4liu, jih023, yip\}@ucsd.edu}}%
\thanks{$^2$Emily K. Funk and Ryan K. Orosco are with the Department of Surgery - Division of Head and Neck Surgery, University of California San Diego, La Jolla, CA 92093 USA. {\tt\small \{ekfunk, rorosco\}@health.ucsd.edu}}}

\begin{document}

\maketitle

\begin{abstract}
Autonomous robotic surgery has seen significant progression over the last decade with the aims of reducing surgeon fatigue, improving procedural consistency, and perhaps one day take over surgery itself.
However, automation has not been applied to the critical surgical task of controlling tissue and blood vessel bleeding--known as hemostasis.
The task of hemostasis covers a spectrum of bleeding sources and a range of blood velocity, trajectory, and volume.
In an extreme case, an un-controlled blood vessel fills the surgical field with flowing blood. 
In this work, we present the first, automated solution for hemostasis through development of a novel probabilistic blood flow detection algorithm and a trajectory generation technique that guides autonomous suction tools towards pooling blood.
The blood flow detection algorithm is tested in both simulated scenes and in a real-life trauma scenario involving a hemorrhage that occurred during thyroidectomy.
The complete solution is tested in a physical lab setting with the da Vinci Research Kit (dVRK) and a simulated surgical cavity for blood to flow through.
The results show that our automated solution has accurate detection, a fast reaction time, and effective removal of the flowing blood.
Therefore, the proposed methods are powerful tools to clearing the surgical field which can be followed by either a surgeon or future robotic automation developments to close the vessel rupture.
\end{abstract}


\begin{figure}[t]
	\centering
	\vspace{2mm}
	\begin{subfigure}{0.23\textwidth}
	\includegraphics[width=1\textwidth]{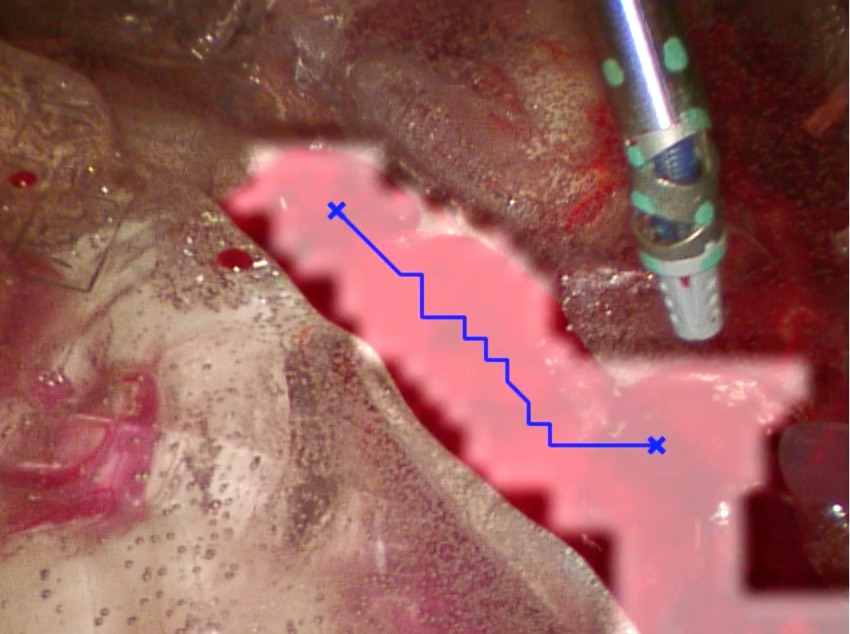}
	\end{subfigure} 
	\begin{subfigure}{0.23\textwidth}
	\includegraphics[width=1\textwidth]{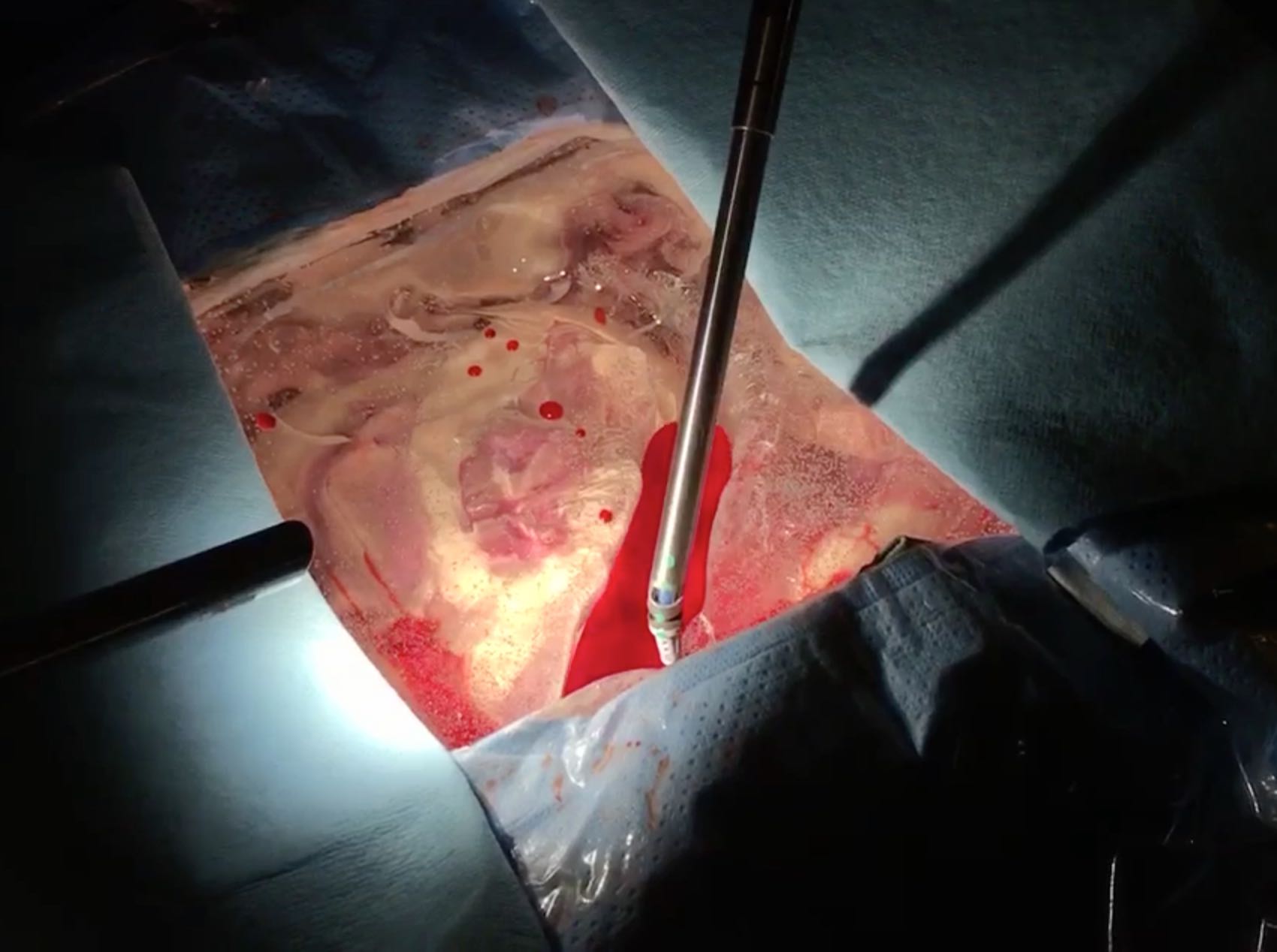}
	\end{subfigure}
	
	\vspace{1mm}
	
    \begin{subfigure}{0.23\textwidth}
	\includegraphics[width=1\textwidth]{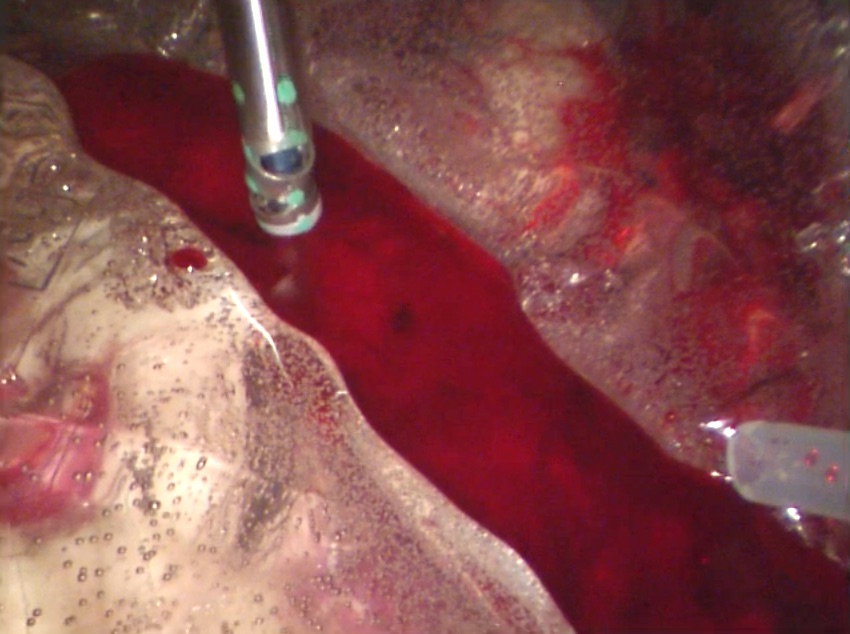}
	\end{subfigure} 
	\begin{subfigure}{0.23\textwidth}
	\includegraphics[width=1\textwidth]{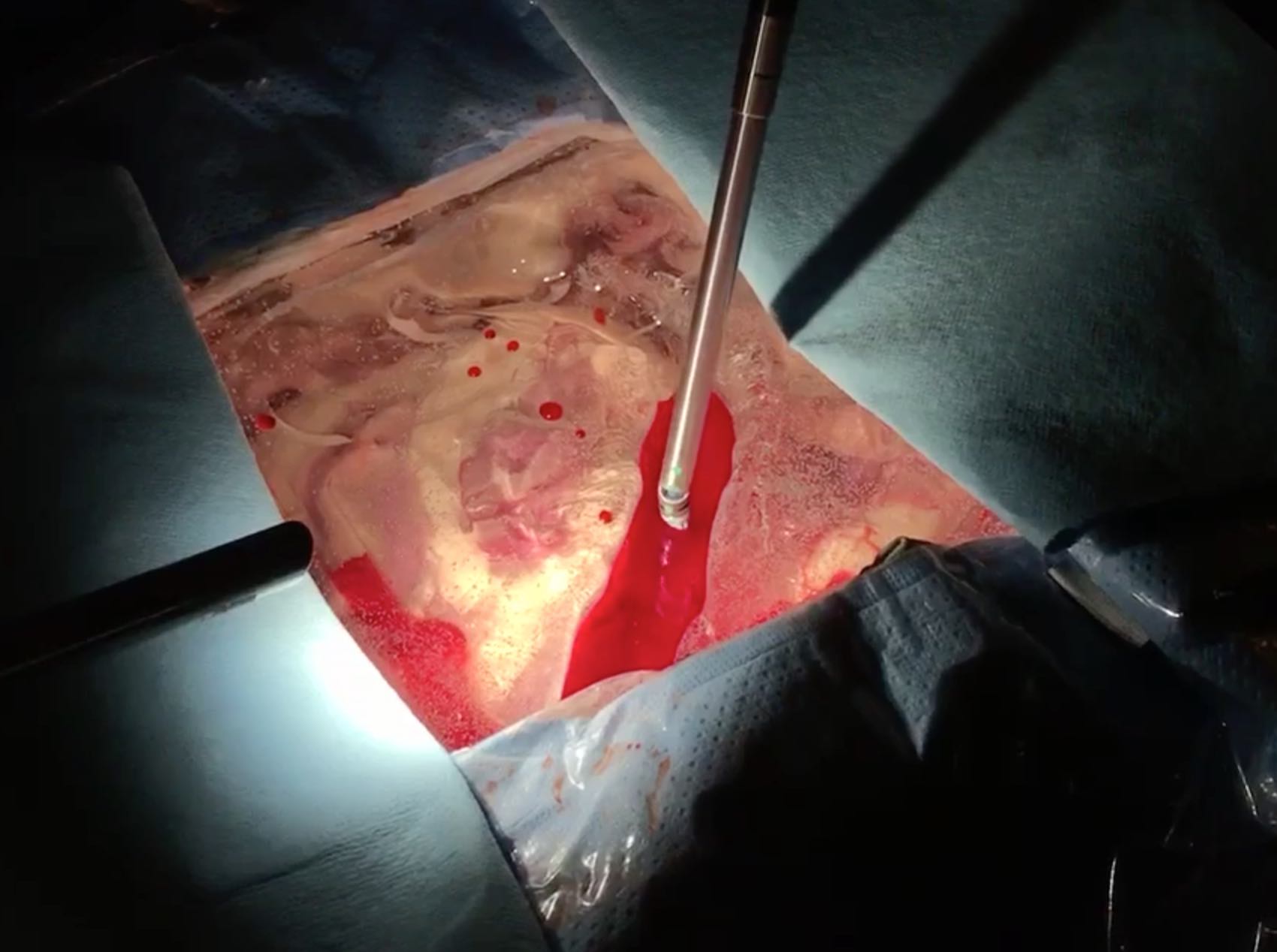}
	\end{subfigure}
		
	\vspace{1mm}
	
    \begin{subfigure}{0.23\textwidth}
	\includegraphics[width=1\textwidth]{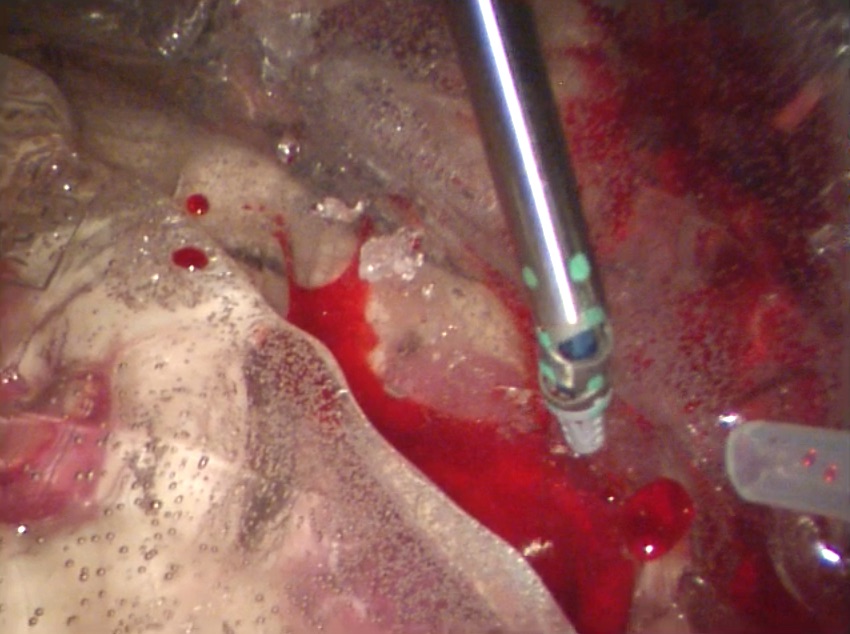}
	\end{subfigure} 
	\begin{subfigure}{0.23\textwidth}
	\includegraphics[width=1\textwidth]{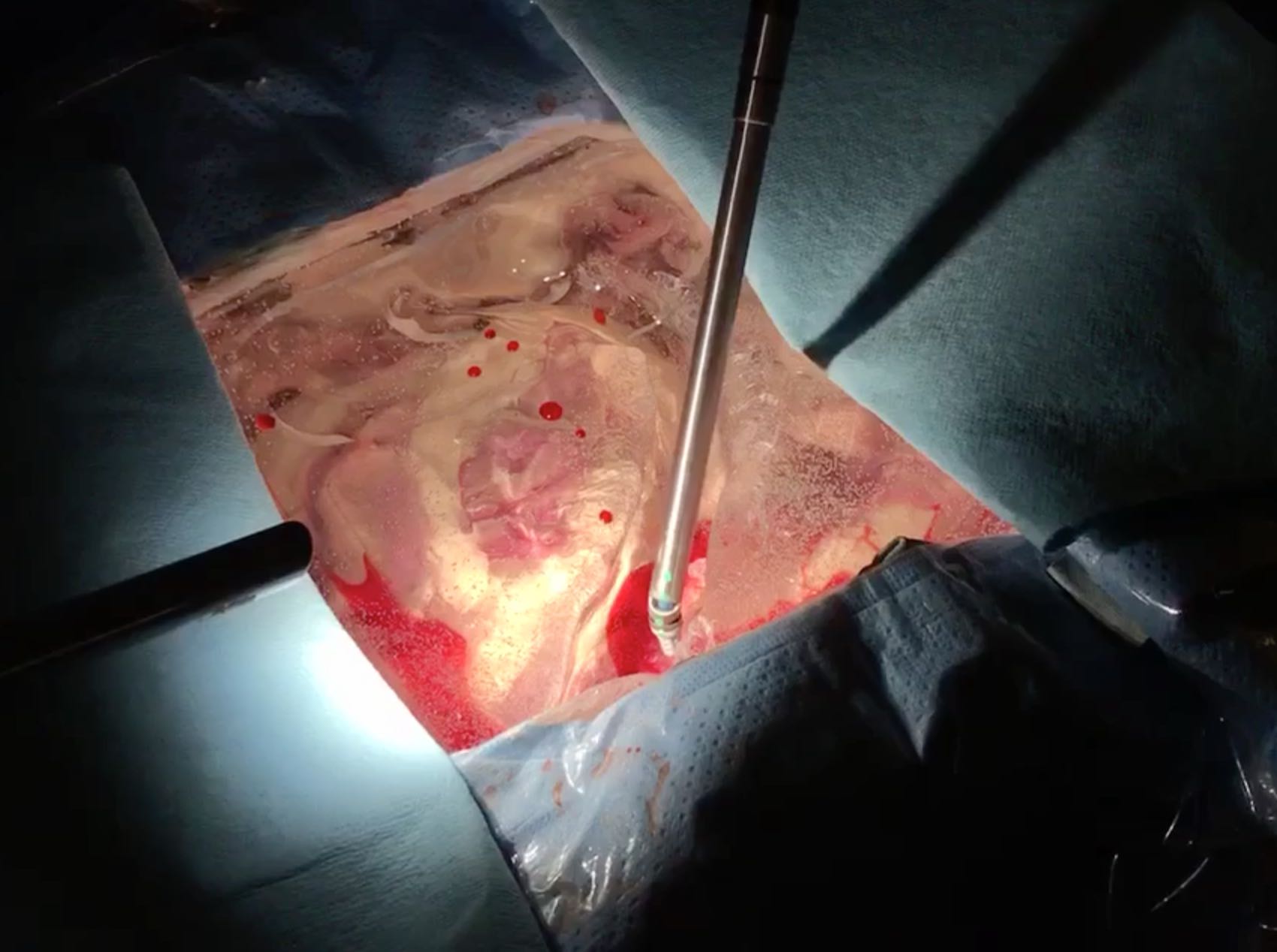}
	\end{subfigure}
	\caption{Autonomous blood suction for cases of trauma in surgery. A robotic suction tool clears the surgical field of flowing blood. The pooling and direction of flowing blood is detected using a temporal, image-based algorithm. This informs a robot trajectory for the suction tool to move upstream towards the source while constrained to stay along the flowing blood.}
    \label{fig:cover_figure}
\end{figure}

\section{Introduction}

Since the deployment of surgical robotic devices such as the da Vinci\textregistered{} Surgical System, efforts to automate surgical tasks have become a popular area of research \cite{yipDasJournal}. The automation of surgery has promised to help reduce surgeon fatigue and improve the procedural consistency between surgeries, and perhaps one day take over surgeries itself to address lack-of-access to timely, life-saving surgery in remote or under-served communities. Success in realizing surgical automation has accelerated in recent years, with improvements in available open-source platforms such as the da Vinci Research Toolkit (dVRK) \cite{kazanzides2014open} and simulators \cite{richter2019open}, coupled with significant advances to data-driven controller synthesis. Successful demonstrations of automated tasks have included cutting \cite{murali2015learning, thananjeyan2017multilateral}, suture needle hand-off \cite{d2018automated}, suturing \cite{sen2016automating, zhong2019dual}, and debridement removal \cite{kehoe2014autonomous}.
Recent developments in perception for surgical robotics helps bridge these autonomous policies in ideal scenarios to realistic, deformable and noisy tissue environments such as the SuPer frameworks \cite{li2020super, lu2020super}. %for tissue manipulation and \cite{alambeigi2018robust} for manipulation of compliant objects). 

While progress in developing autonomous surgical tasks has leapt forward, a key area that has been given little attention are reactive policies to traumatic events, such as hemostasis.
Hemostasis describes a state of the surgical field that is fulfilled when there is no site of active bleeding and the tissues are unobstructed by blood.
The bleeding can originate from a visible or macroscopic blood vessel (artery or vein), or from the microvasculature and capillary network within tissues.
Unlike previously automated tasks that occur in a more predictable cadence within a procedure, bleeding can be unpredictable which necessitates hemostasis maneuvers at any time during any surgery.
Surgical manipulation of any type--suction, grasping, retraction, cutting, dissection--can immediately cause bleeding.
Bleeding can also occur in a delayed manner, for example if a vessel is not definitively sealed, it can rupture spontaneously without direct contact.
If surgical automation is ever to be deployed, reactive policies will be required to handle these traumatic scenarios.

This work specifically addresses the problem of small and medium vessel ruptures.
Overall, there are four distinct stages in hemostasis of this scenario: (1) clearing the surgical field of blood; (2) identification of the bleeding source (vessel rupture); (3) grasping the identified vessel to temporarily stop bleeding; (4) closing the vessel rupture, usually with an energy device, clip, or suture.
Each stage requires a complex set of manipulation skills as well as perception algorithms that make it non-trivial to implement.

In this paper, we describe an automated solution to the first task, clearing the surgical field, as shown in Fig. \ref{fig:cover_figure}.
This task involves first recognizing blood in a scene, then tracking blood flow temporally, and finally prescribing a real-time trajectory generation strategy that will intelligently control a suction tool to siphon the blood to efficiently suction it.
To this end, we present the following novel contributions:
\begin{enumerate}
    \item The first complete automated solution for clearing the surgical field of flowing blood from a ruptured vessel using a robotic suction tool,
    \item a novel blood flow detection and tracking method by utilizing temporal visual information,
    \item and a trajectory generation technique from blood regions in the image frame for a surgical suction tool to follow and clear the surgical field.
\end{enumerate}
The blood flow detection and tracking method is tested within various simulated scenes as well as a real-life case involving a vessel rupture during thyroidectomy. 
The complete solution is tested in a lab setting with the da Vinci Research Kit (dVRK) \cite{kazanzides2014open} and a simulated surgical cavity for blood to flow through and collect in. 
The results from the experiments show the effectiveness of the blood flow tracking and surgical suction tool trajectory generation developed in this work.

\section{Related Works}

Previous work on blood detection largely is from the context of Wireless Capsule Endoscopy (WCE) where image processing for detections is used to speed up clinician workflow \cite{liedlgruber2011computer}.
The typical approach to blood detection in WCE is to classify either on a pixel level or using patch-based methods \cite{fu2013computer}.
The feature space used for classification is either direct Red, Green, Blue (RGB) \cite{liu2009obscure} channels or the transformed Hue, Saturation, Value (HSV) channels \cite{jung2008active}. %For RGB \cite{penna2009technique}
To efficiently process these color spaces, techniques such as support vector machines \cite{okamoto2019real}, chromium moments combined with binary pattern textures \cite{li2009computer}, and neural networks \cite{pan2011bleeding, fu2011bleeding} have been demonstrated. 
While these methods are robust to small individual lesions, in a surgical scene there can be stains from previous ruptures and larger amounts of blood flow that make the problem of blood detection and, specifically, tracking, a more challenging and complex problem. 

There has been previous research on robots interacting with liquids in the act of pouring \cite{mottaghi2017see}. %\cite{rozo2013force}
However, these methods cannot be applied to a surgical setting since they are limited to specific objects for pouring and capturing liquids.
Schenck and Fox applied deep neural networks to detect fluids \cite{schenck2018perceiving} that can be combined with differential fluid dynamics to reconstruct 3D information \cite{schenck2018spnets}.
This detection method however requires labelled real-world data which are challenging to collect in a surgical context.
Yamaguchi and Atkeson instead used the heuristic of optical flow to detect moving fluid \cite{yamaguchi2016stereo}.
The work presented here also uses optical flow to detect blood flow.
However, instead of the classical method used by Yamaguchi and Atkeson, we applied a deep learning technique for improved performance of optical flow estimation in a surgical environment and fused the detections temporally with a novel temporal filter.

% %Trick to add spacing above algorithm, otherwise algorithm fails margin test by IEEE
\begin{figure}[t]
    \vspace{-2.5mm}
\end{figure}

\begin{algorithm}[t]
    % \BlankLine
    \caption{Blood Suction}
    \label{alg}
 Initialize image frame probability map of blood: $\mathbf{P}_0$ \\
 Initialize optical flow CNN with $l$ image frame inputs \\
 Initialize counts of being blood: $\mathbf{C}_t  \leftarrow \mathbf{0}$\\
 \While{new image, $\mathbf{I}_t$, arrives}{
  // Blood Region Detection\\
  $\mathbf{O}_t  \leftarrow \text{getOpticalFlowCNN}(\mathbf{I}_t, \dots, \mathbf{I}_{t-l-1})$\\
  \For{pixel $p_{t} \in \mathbf{P}_{t}$}{
  // Detect: \\
    \eIf{$||\mathbf{O}_t(p)|| > \gamma_o$}{$z^p_t  \leftarrow \text{b}$}
    {$z^p_t  \leftarrow \overline{\text{b}}$}
    Predict: $P(p_t = \text{b} | z^p_{1:t-1})$ with (\ref{eq:predict_step}) and $\mathbf{P}_{t-1}$\\
    Update: $P(p_t  = \text{b} | z^p_{1:t})$ with (\ref{eq:update_step}) and $z^p_t$\\
  }
  $\mathbf{B}_t$ is set to the largest connected region in binary mask $\mathbf{P}_t > 0.5$ \\
  \If{ $size(\mathbf{B}_t) <  \gamma_B$}{
        continue\\
  }
  
  // Suction Trajectory Generation\\
  Update Blood Count: $\mathbf{C}_t  \leftarrow \mathbf{C}_{t-1} + \mathbf{B}_t$ \\
  End point: $e_t  \leftarrow \underset{pixel}{\mathrm{argmax }} \text{ } \mathbf{C}_t \wedge \mathbf{B}_t$ \\
  Start point: $s_t  \leftarrow \underset{pixel}{\mathrm{argmin}} \text{ } \mathbf{C}_t \wedge \mathbf{B}_t$ \\
  $\mathbf{T}_t  \leftarrow $ generateTrajectoryForSuction($s_t$, $e_t$, $\mathbf{B}_t$)\\
  \If{length$(\mathbf{T}_t) > \gamma_T$}{
    break\\
  }
 }
 executeTrajectory($\mathbf{T}_t$)
\end{algorithm}

\section{Methods} 

An overview of the proposed algorithms for blood region estimation and trajectory generation for an autonomous suction tool on a surgical robot is provided in Algorithm \ref{alg}.
At a high level, the blood region is estimated by updating a probability map on the scene, which describes the probability of each pixel in the image frame being blood or not.
From the probability map, the blood region is extracted.
A trajectory is then generated for the suction tool to follow in order to clear the surgical field from blood.
The trajectory is constrained to stay within the blood region to maximize the blood removed.

\begin{figure}[t]
	\centering
	\vspace{2mm}
	\begin{subfigure}{0.2225\textwidth}	\includegraphics[trim=10.5cm 1.25cm 10.5cm 0.5cm, clip, width=1\textwidth]{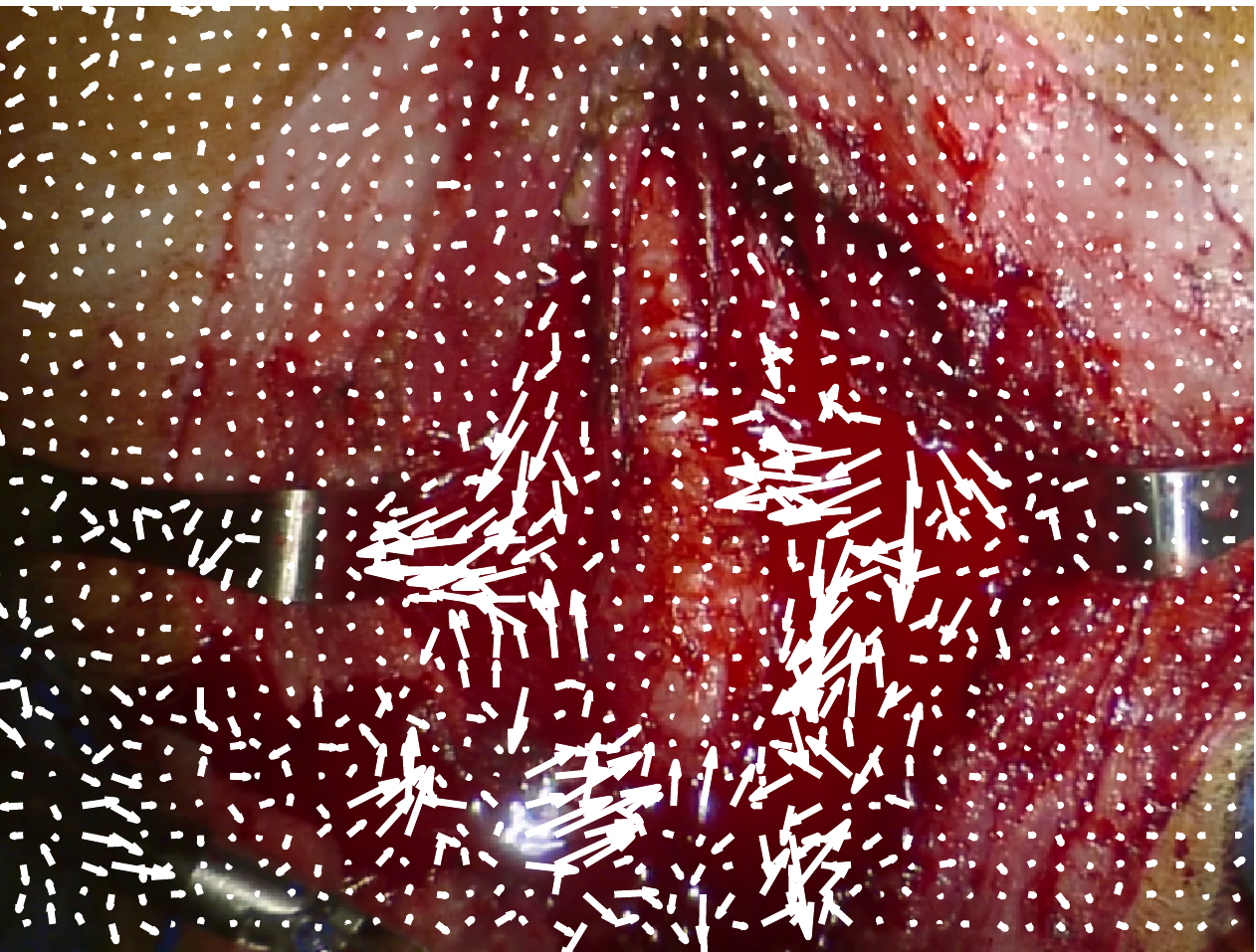}
	\end{subfigure} 
	\begin{subfigure}{0.255\textwidth}	\includegraphics[trim=12.37cm 6.56cm 11.5cm 2.3cm, clip, width=1\textwidth]{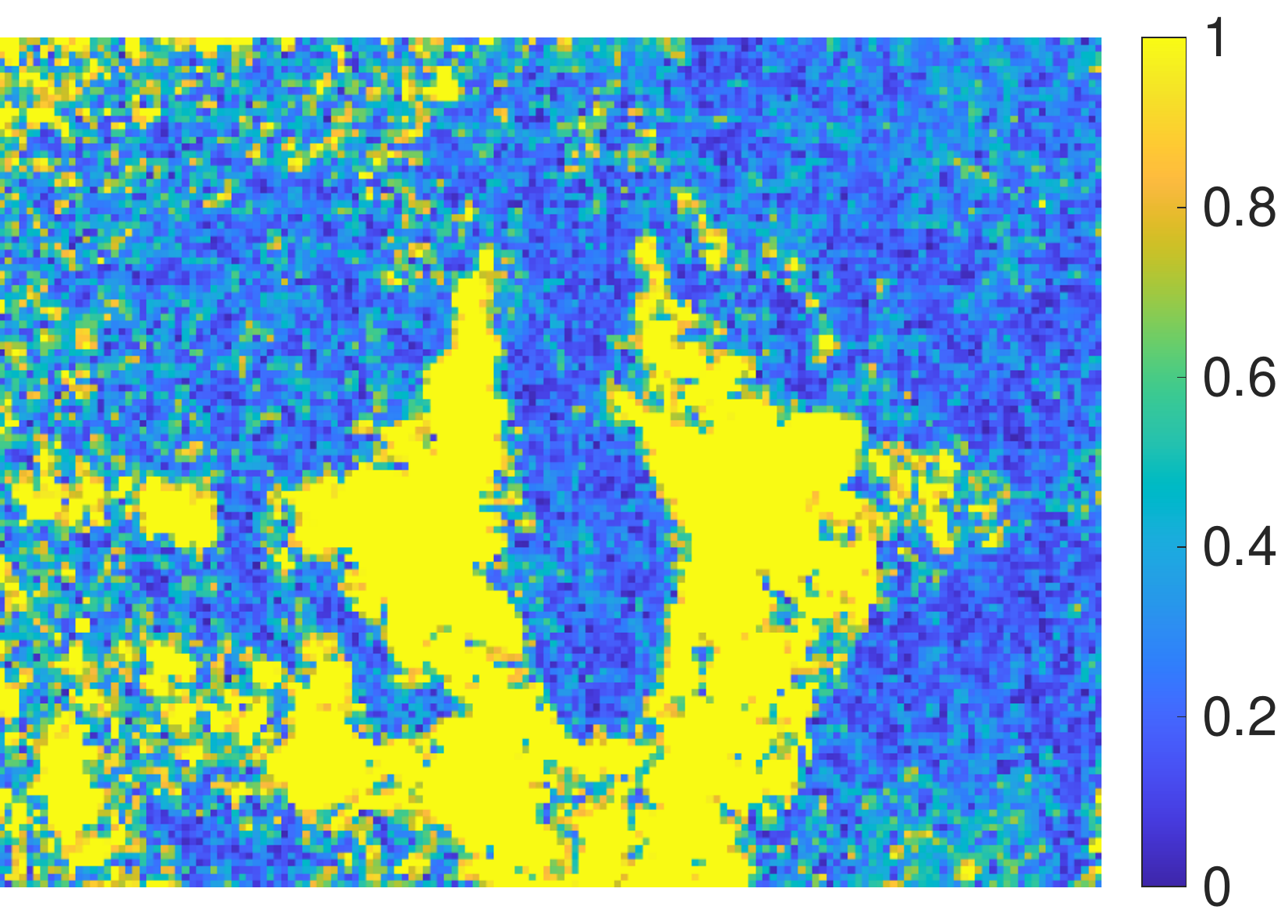}
	\end{subfigure}
	\caption{The above images are generated by optical flow estimation from an in-vivo surgical scene. The left image shows the vectors of estimated image motion from optical flow, and the right image is a normalized heatmap of their magnitudes. Notice that the magnitude of optical flow detects the regions of blood flow well, while the orientation gives inconsistent information about the flow.
	}
    \label{fig:optical_flow_example_image}
\end{figure}

\subsection{Detecting Flowing Blood in Image Frame}

Optical flow is chosen to detect flowing blood because it extracts information about all moving objects in the scene. 
In a surgical scene, the main motion comes from surgical tools and flowing fluids.
Another source of motion in robotic surgery comes from a moving camera, but for robotic surgery the camera remains stationary when work is being done in a scene and its position is reset only to change the field of view.
We therefore consider only stationary scenes.
To mask instrument motion from the scene, previously developed methods can be applied to effectively segment and remove pixels attributed to surgical tools from image \cite{li2020super, lu2020super, allan20192017}.

To estimate optical flow, a pretrained convolutional neural network (CNN) is used \cite{teney2016learning}.
A deep learning strategy is used instead of traditional methods such as Lucas-Kanade \cite{lucas1981iterative} (used in previous work in robot pouring \cite{yamaguchi2016stereo}) 
% The appeal of the CNN approach is to use learned features to identify local motion cues rather than scene-level interpretation.
as traditional optical flow approaches utilize brightness constancy constraint assumption, and this assumption is not valid in endoscopic procedures due to irregular lighting.
Meanwhile, the proposed architecture by Teney and Herbert \cite{teney2016learning} is able to extract motion from learned features that are invariant to textures, brightness, and contrast, which is ideal for detecting flowing blood from an endoscope.

Similar to previous work in robot pouring \cite{yamaguchi2016stereo}, we also found experimentally that the magnitude of optical flow is a good signal for detecting fluid motion while the orientation is not.
An example of the processed image is shown in Fig. \ref{fig:optical_flow_example_image} comparing the RGB image to the amplitude map for optical flow.

Consider the magnitude of optical flow at pixel $p$.
Let $z^p_t$ be the random variable describing the detection of blood at pixel $p_t$ at time $t$.
The detection is modelled where blood is detected, $z^p_t = \text{b}$, if the magnitude of optical flow at pixel $p_t$ is greater than a threshold, $\gamma_o$.
The inverse is also set, so no blood is detected, $z^p_t = \overline{\text{b}}$, if the magnitude at pixel $p_t$ is less than $\gamma_o$.
Hence the probability model for these detections can be simply written as:
\begin{align}
        P( z^p_t  = \text{b} \mid p_t = \text{b})  && P(z^p_t = \text{b} \mid  p_t = \overline{\text{b}})
        \label{eq:detection_probabilities_blood}
\end{align}
which describes an observation model for the hidden state $p_t \in \{ \text{b}, \overline{\text{b}} \}$.

\subsection{Temporal Filtering for Blood Region Detection}

Although the magnitudes of optical flow provide a good initial estimate for blood detection, they are nevertheless noisy and require filtering.
Therefore, a temporal filter is based on a Hidden Markov Model (HMM) is proposed to fuse independent measurements of the pixel labels over time.
The HMM tracks the discrete states for $p_t$ using the observation models in (\ref{eq:detection_probabilities_blood}).
Let the following be a transition probability for a pixel $p_t$ be
\begin{equation}
    P(p_{t+1} = \text{b} \mid p_t = \text{b}),
    \label{eq:motion_model_blood_to_blood}
\end{equation}
which models the probability that if a pixel is already blood it will continue being blood.
In the case of blood vessel ruptures, this should be set close to 1 since the vessel rupture will not stop emitting blood until it has been closed.
For the transition probabilities where the pixel is not blood at time $t$, an additional parameter, $k^p_t$, is introduced to the model:
\begin{align}
     \label{eq:motion_model_no_blood_to_blood}
     P(p_{t+1} = \text{b} \mid p_t = \overline{\text{b}}, k^p_t = \text{b}) \\ P(p_{t+1} = \text{b} \mid p_t = \overline{\text{b}}, k^p_t = \overline{\text{b}}),
     \label{eq:motion_model_no_no_blood_to_blood}
\end{align}
where $k^p_t$ describes the state of the neighboring pixels of $p_t$.
This is modelled as the resulting Boolean-OR operation ($\vee$) on the states of the neighboring pixels:
\begin{equation}
    k^p_t = \bigVee_{q^i \in \mathbf{A}_p} q^i_t
\end{equation}
where $\mathbf{A}_p$ is the set of neighboring pixels to $p_t$. 
Therefore, the model from (\ref{eq:motion_model_no_blood_to_blood}) is capturing the flow of the blood and (\ref{eq:motion_model_no_no_blood_to_blood}) is describing the probability a blood source starts at pixel $p_t$.
To appropriately describe these processes in this case of blood vessel ruptures, (\ref{eq:motion_model_no_blood_to_blood}) should be set less than (\ref{eq:motion_model_blood_to_blood}) and (\ref{eq:motion_model_no_no_blood_to_blood}) close to 0.
% Intuitively, it can be seen that increasing the size of the neighborhood $|\mathbf{A}_p|$ is similar to increasing the pixel velocity of blood flow. 

The temporal filter is designed to estimate the posterior probability of the state $p_t$ using transition probabilities and observation models.
This is done using a predict and update step after every detection.
The predict step can be calculated as:
\begin{multline}
        P(p_{t+1} \mid z^p_{1:t}) = P(p_{t+1} \mid p_t = \text{b}) P(p_t = \text{b} \mid z^p_{1:t}) + \\
        \sum \limits_{k^p_t \in \{\text{b}, \overline{\text{b}}\}} P(p_{t+1} \mid p_t = \overline{\text{b}}, k^p_t) P(p_t = \overline{\text{b}}, k^p_t \mid z^p_{1:t})
\end{multline}
and the update step is computed:
\begin{equation}
    P(p_{t+1} \mid z^p_{1:t+1}) \propto  P(z_{t+1} \mid p_{t+1}) P(p_{t+1} \mid z^p_{1:t})
    \label{eq:update_step}
\end{equation}
However, the predict expression has the joint probability of $p_t$ and $k^p_t$.
Explicit estimation for this joint probability would be computationally intractable, so each pixel's probability of being blood is approximated to be independent of all others at time $t$.
With this simplification, the predict step can be rewritten as:
\begin{multline}
    P(p_{t+1} \mid z^p_{1:t}) = P(p_{t+1} \mid p_t = \text{b}) P(p_t = \text{b} \mid z_{1:t}) + \\
    \sum \limits_{k^p_t \in \{\text{b}, \overline{\text{b}}\}} P(p_{t+1} \mid p_t = \overline{\text{b}}, k^p_t)  P(k^p_t \mid z^p_{1:t}) P(p_t = \overline{\text{b}}\mid z^p_{1:t}) 
    \label{eq:predict_step}
\end{multline}
and an expression can be found for $P(k^p_t \mid z^p_{1:t})$ using the inclusion-exclusion principle:
\begin{equation}
    P(k^p_t \mid z^p_{1:t}) = \sum \limits_{j = 1}^{|\mathbf{A}_p|}  (-1)^{j-1}  \sum \limits_{\substack{\mathbf{J} \subseteq  \mathbf{A}_p\\|\mathbf{J}| = j}} \prod \limits_{q^i \in \mathbf{J}} P( q^i_t \mid z^p_{1:t}) 
\end{equation}
This results in the ability to compute and track the probabilities of each pixel being blood using (\ref{eq:predict_step}) and (\ref{eq:update_step}) after every detection.

\begin{figure}[t]
    \vspace{-3mm}
\end{figure}

\begin{algorithm}[t]
    \caption{Trajectory Generation for Suction}
     \label{alg:traj}
    \SetKwInOut{Input}{Input}
    \SetKwInOut{Output}{Output}
  \Input{Start pixel, $s_t$, end pixel, $e_t$, and mask of blood region $\mathbf{B}_t$}
   \Output{Trajectory $\mathbf{T}_t$}
  Initialize clearance reward $\mathbf{R}  \leftarrow \mathbf{0}$\\
  Initialize temporary eroded blood region $\mathbf{E}  \leftarrow \mathbf{B}_t$\\
  Initialize $i \leftarrow 0$ \\
  \While{$\mathbf{E} \neq \mathbf{0}$ and $i < \gamma_r$}{
        $\mathbf{E}  \leftarrow \text{erode}(\mathbf{E})$ \\
        $\mathbf{R}  \leftarrow \mathbf{R} + r\mathbf{E}$\\
        $i \leftarrow i + 1$
  }
 Initialize pixels cost to go to goal: $\mathbf{D}  \leftarrow \mathbf{\infty}$\\
 Initialize visited map: $\mathbf{V}  \leftarrow ! \mathbf{B}_t$ \\
 Initialize parents of nodes: $\mathbf{K}  \leftarrow 
 \text{UNDEFINED}$\\
 $\mathbf{D}(s_t)  \leftarrow 0$\\
 \While{ $\mathbf{V} \neq \mathbf{1}$ }{
    $u  \leftarrow \underset{pixel}{\mathrm{argmin}} \text{ } \mathbf{D}$ \\
    $\mathbf{V}(u)  \leftarrow 1$\\
    \If{$u = e_t$}{
        break\\
    }
    \For{pixel $q$ neighboring $u$ and $\mathbf{V}(q) \neq 1$}{
        \If{$\mathbf{D}(q) > ||q - u|| - \mathbf{R}(q)$}{
        $\mathbf{D}(q)  \leftarrow ||q - u|| - \mathbf{R}(q)$\\
        $\mathbf{K}(q)  \leftarrow u$\\
        }
    }
 }
 Initialize trajectory for output: $\mathbf{T}_t  \leftarrow $ [ ]\\
 Initialize parent traversal node: $u  \leftarrow e_t$\\
 \While{$u \neq s_t$}{
    Insert $u$ at beginning of $\mathbf{T}_t$\\
    $u  \leftarrow \mathbf{K}(p)$ \\
 }
 \Return{ $\mathbf{T}_t$ } 
\end{algorithm}

To find the region of blood on the image frame, a mask is generated where all pixels with a posterior probability greater than 0.5 is set to 1, and the rest are set to 0.
Then dilation and erosion morphological operations are applied once to reduce noise on the mask.
Finally, the largest connected region of the mask is considered the region with blood flowing if its size is greater than a threshold of $\gamma_B$.
This threshold keeps a detection from occurring when there is no actual blood flowing.

\begin{figure}[t]
	\centering
	\vspace{2mm}
	\begin{subfigure}{0.47\textwidth}
    \includegraphics[trim=4cm 1.5cm 4cm 4cm, clip, width=1\textwidth]{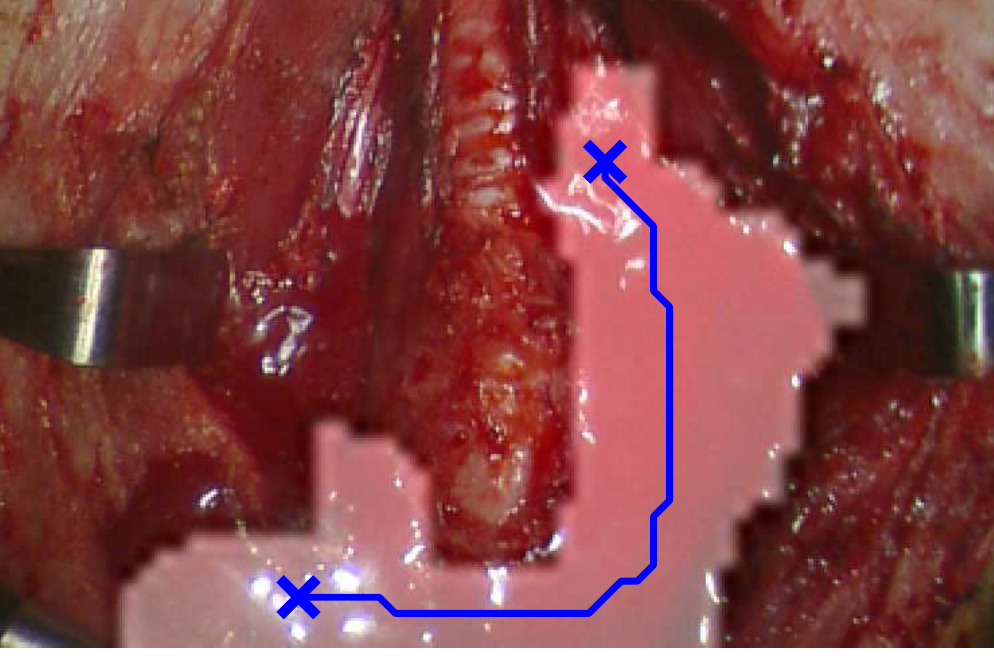}
	\end{subfigure}
	\begin{subfigure}{0.215\textwidth}
	\includegraphics[trim=1.9cm 1.8cm 1.5cm 1.65cm, clip, width=1\textwidth]{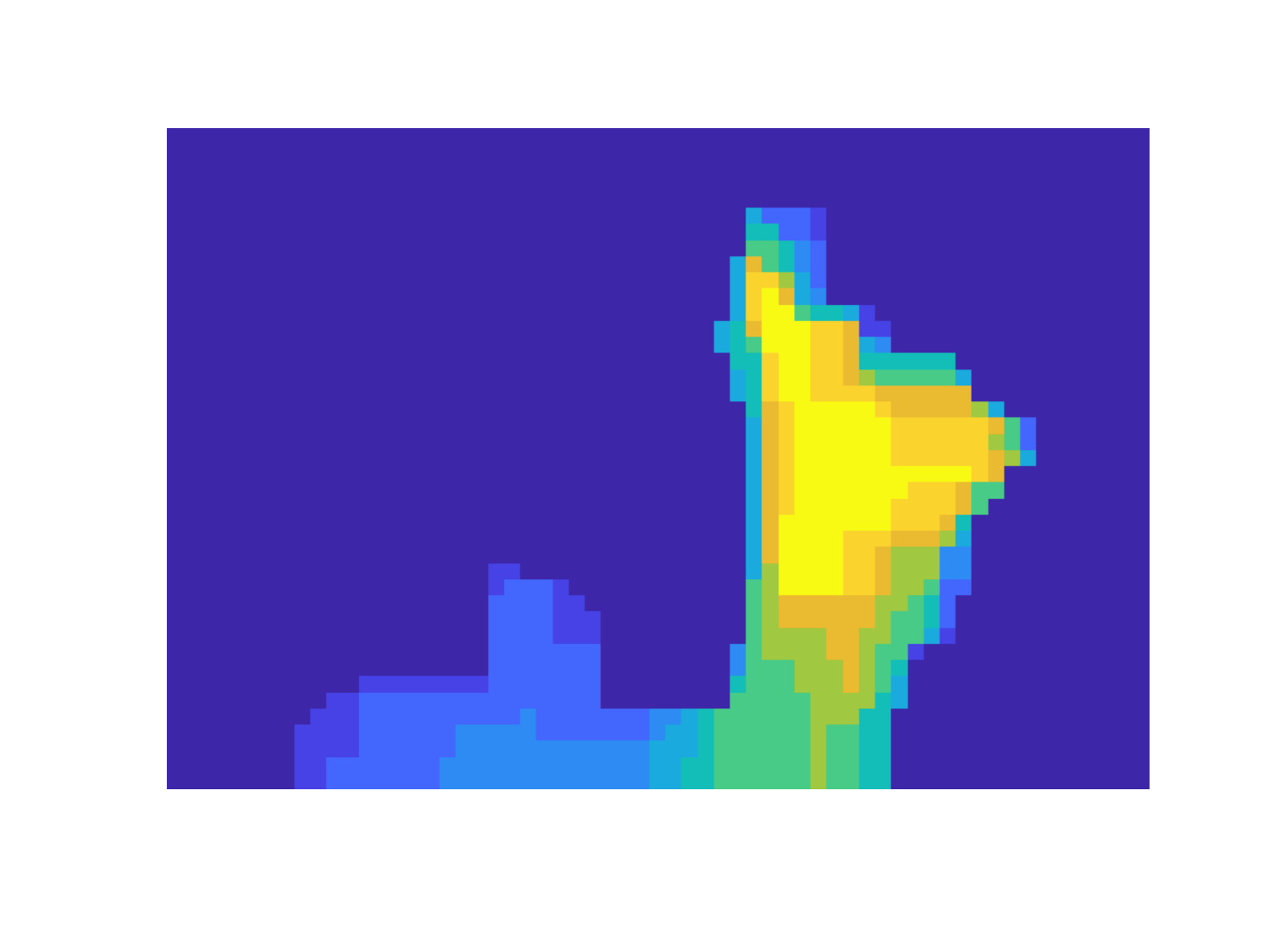}
	\end{subfigure} 
	\begin{subfigure}{0.25\textwidth}
	\includegraphics[trim=1.8cm 2.0cm 0.9cm 1.6cm, clip, width=1\textwidth]{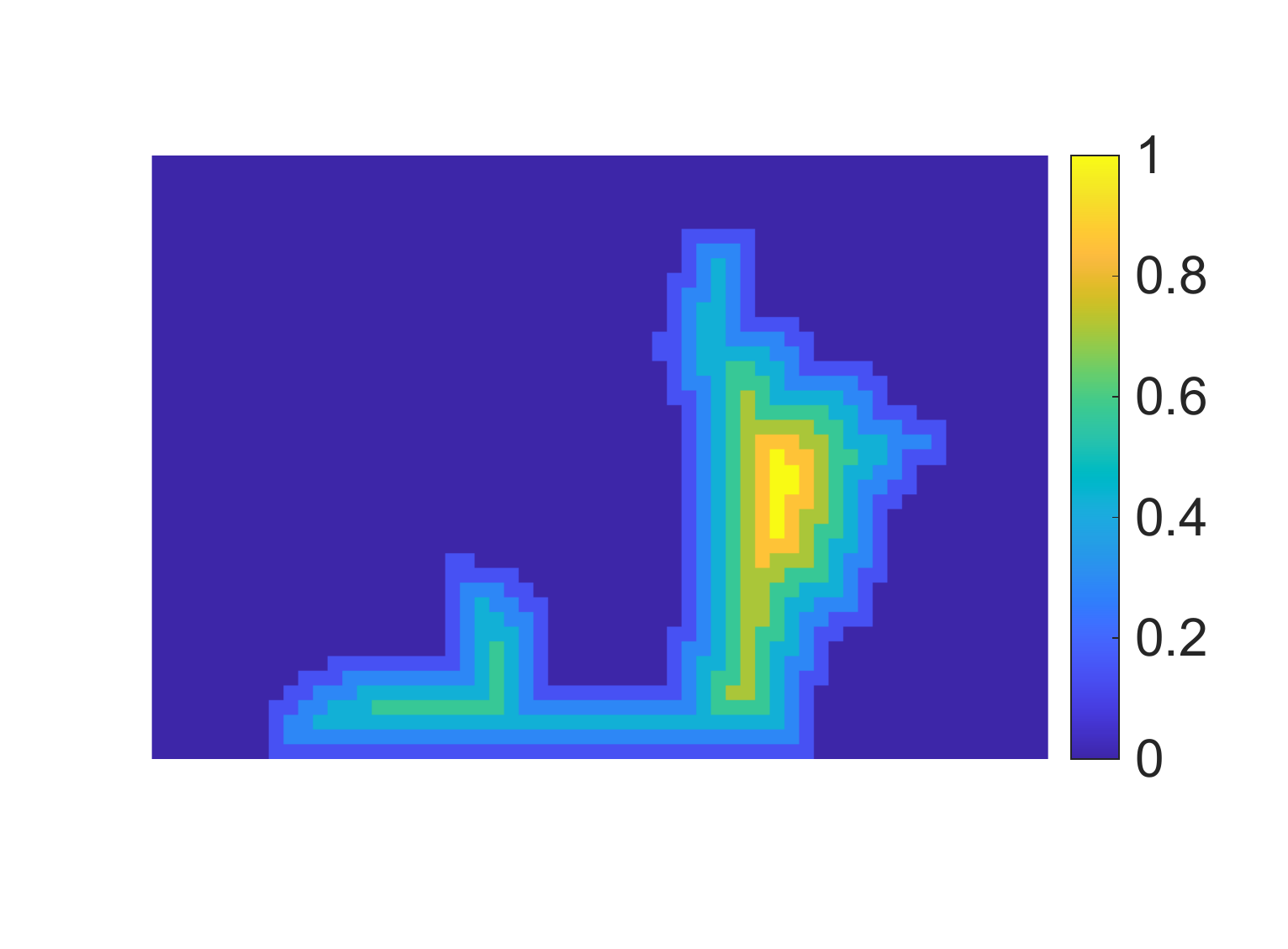}
	\end{subfigure} 
	\caption{Example of trajectory generation for blood suction from in-vivo scene. In the top figure, the white highlighted region is the detected blood region and the blue path is the generated trajectory. The starting and stopping points are chosen from the newest and oldest regions of the detected blood respectively, and a heatmap of the normalized detected blood age is shown in the bottom left figure. The generated path uses an additional clearance reward, which is shown in the bottom right figure, to reward paths that stay centered along the blood stream.}
    \label{fig:trajectory_gen_example}
\end{figure}

\begin{figure*}[t]
    \centering
    \vspace{2mm}
    \begin{subfigure}{0.16\textwidth}
	\includegraphics[trim=0cm 0cm 0cm 3cm, clip, width=1\textwidth]{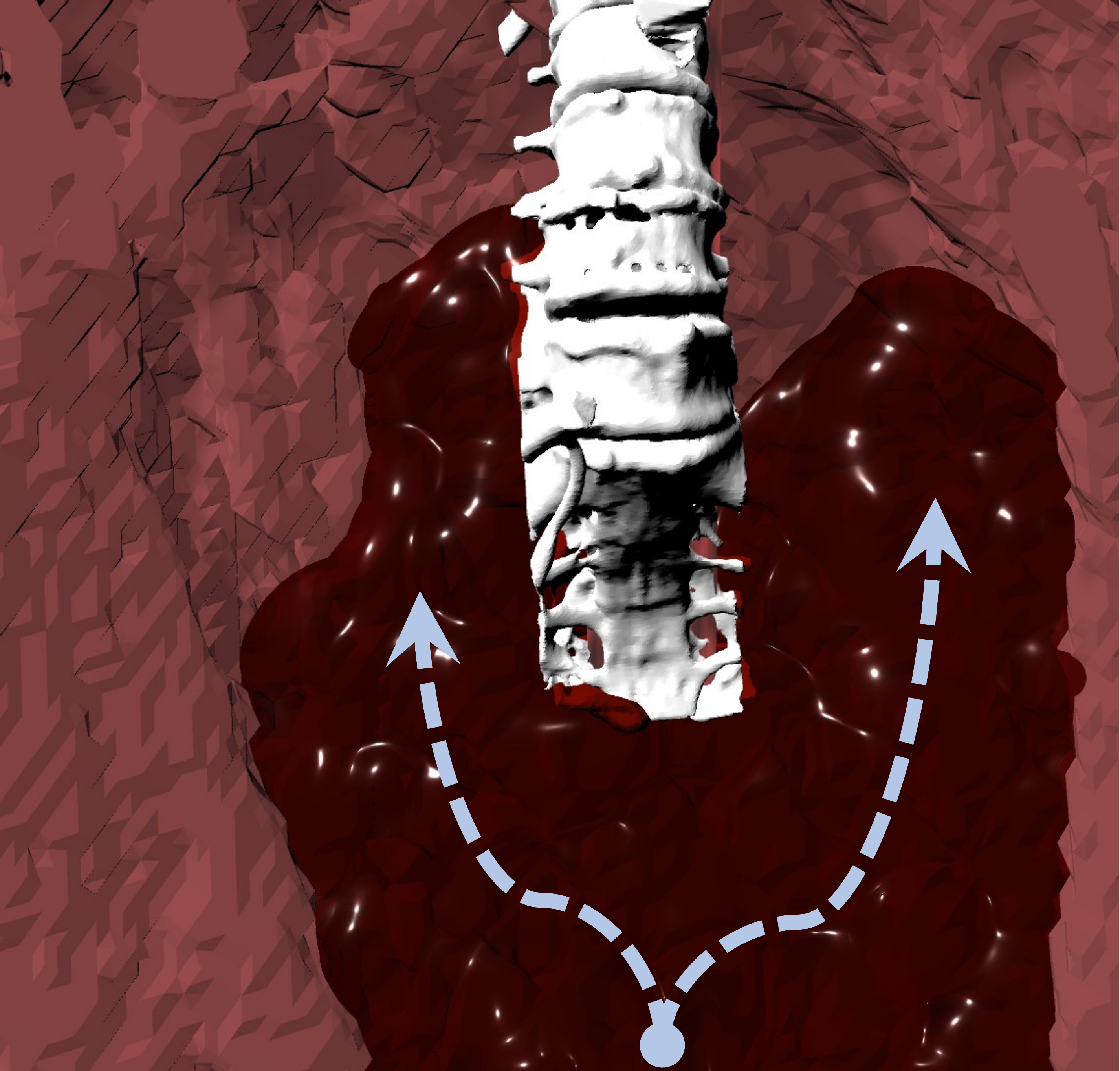}
	\end{subfigure}
    \begin{subfigure}{0.16\textwidth}
	\includegraphics[width=1\textwidth]{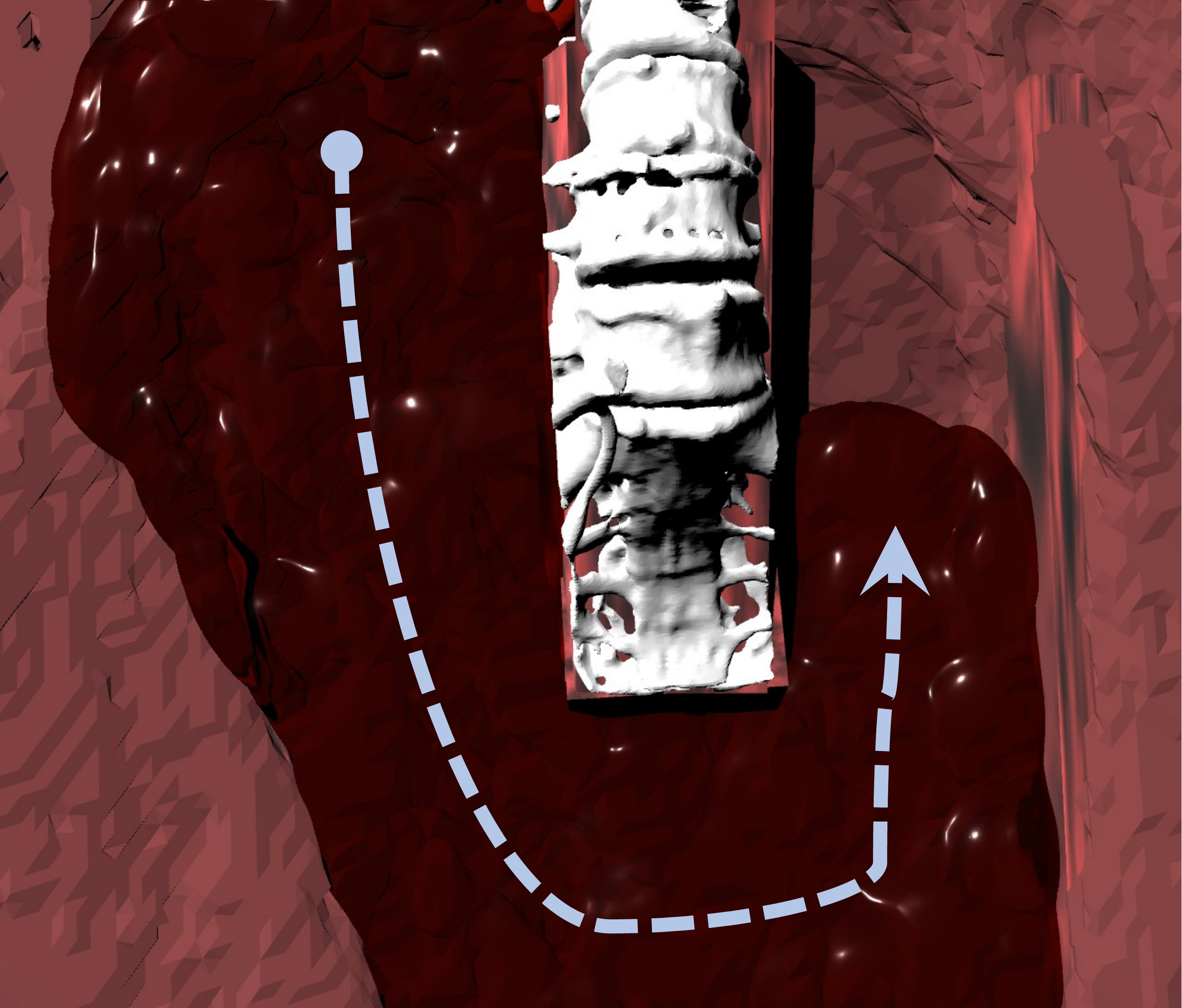}
	\end{subfigure}
	\begin{subfigure}{0.16\textwidth}
	\includegraphics[width=1\textwidth]{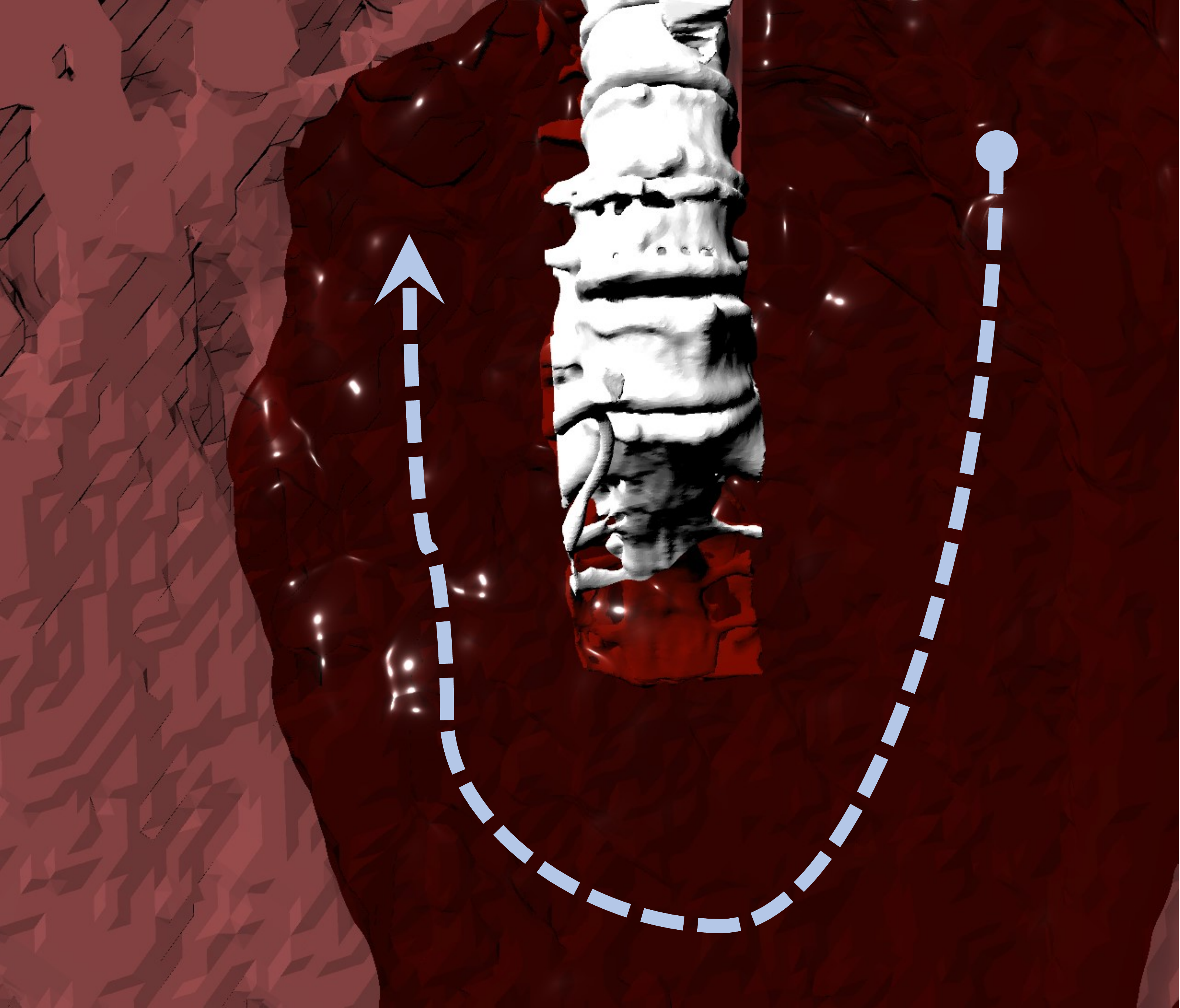}
	\end{subfigure}
	\begin{subfigure}{0.16\textwidth}
	\includegraphics[width=1\textwidth]{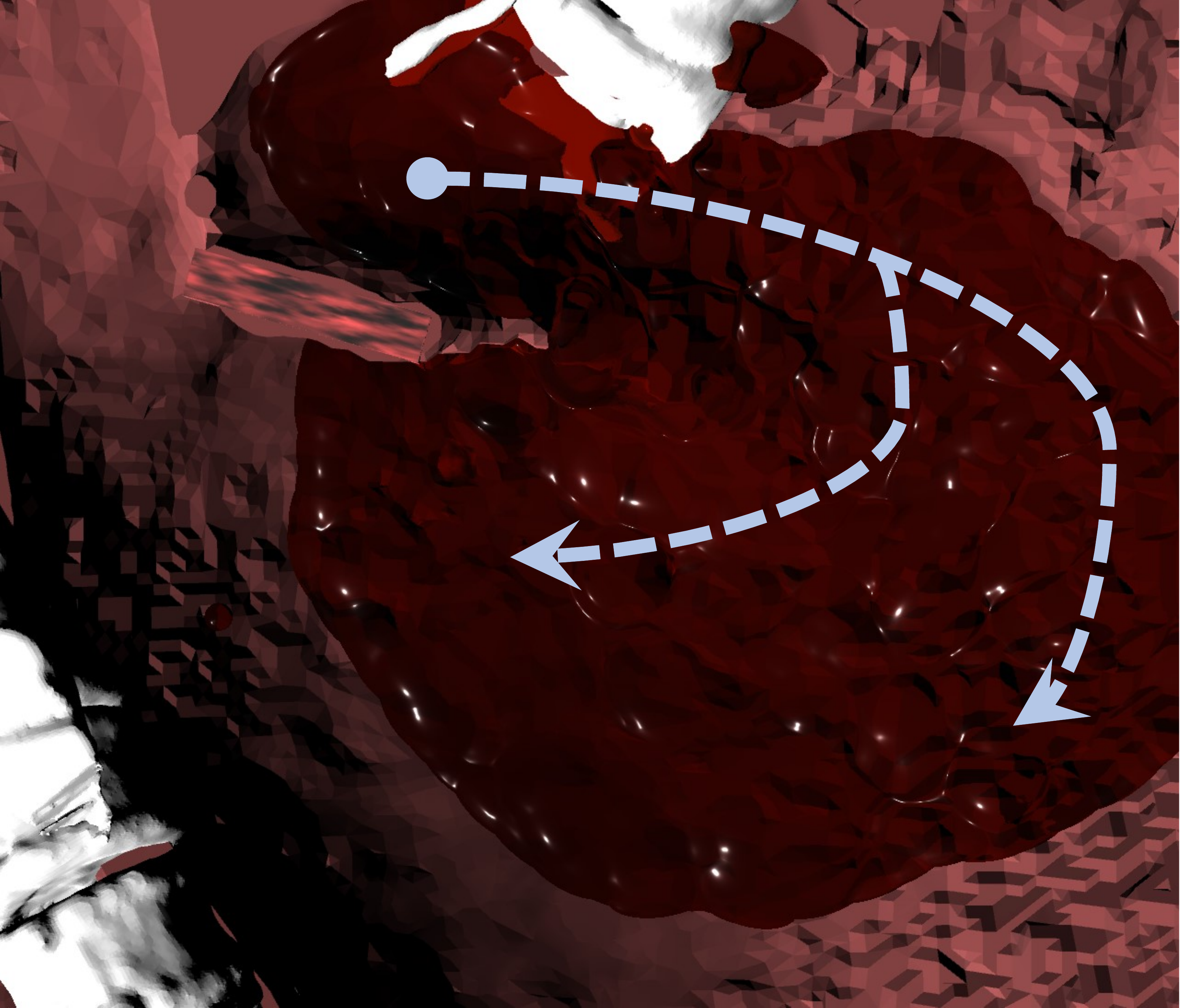}
	\end{subfigure} 
	\begin{subfigure}{0.16\textwidth}
	\includegraphics[width=1\textwidth]{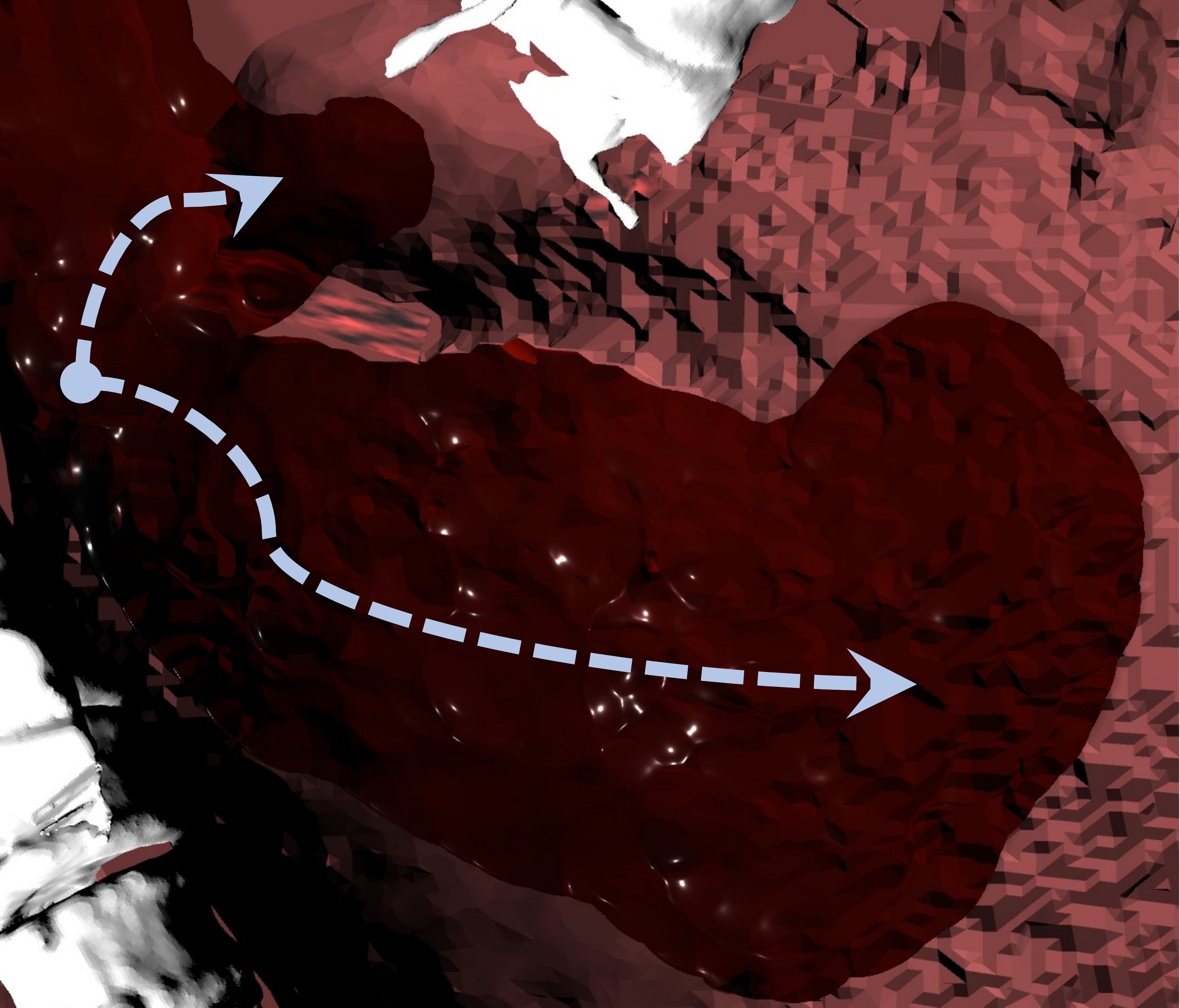}
	\end{subfigure} 
	\begin{subfigure}{0.16\textwidth}
	\includegraphics[width=1\textwidth]{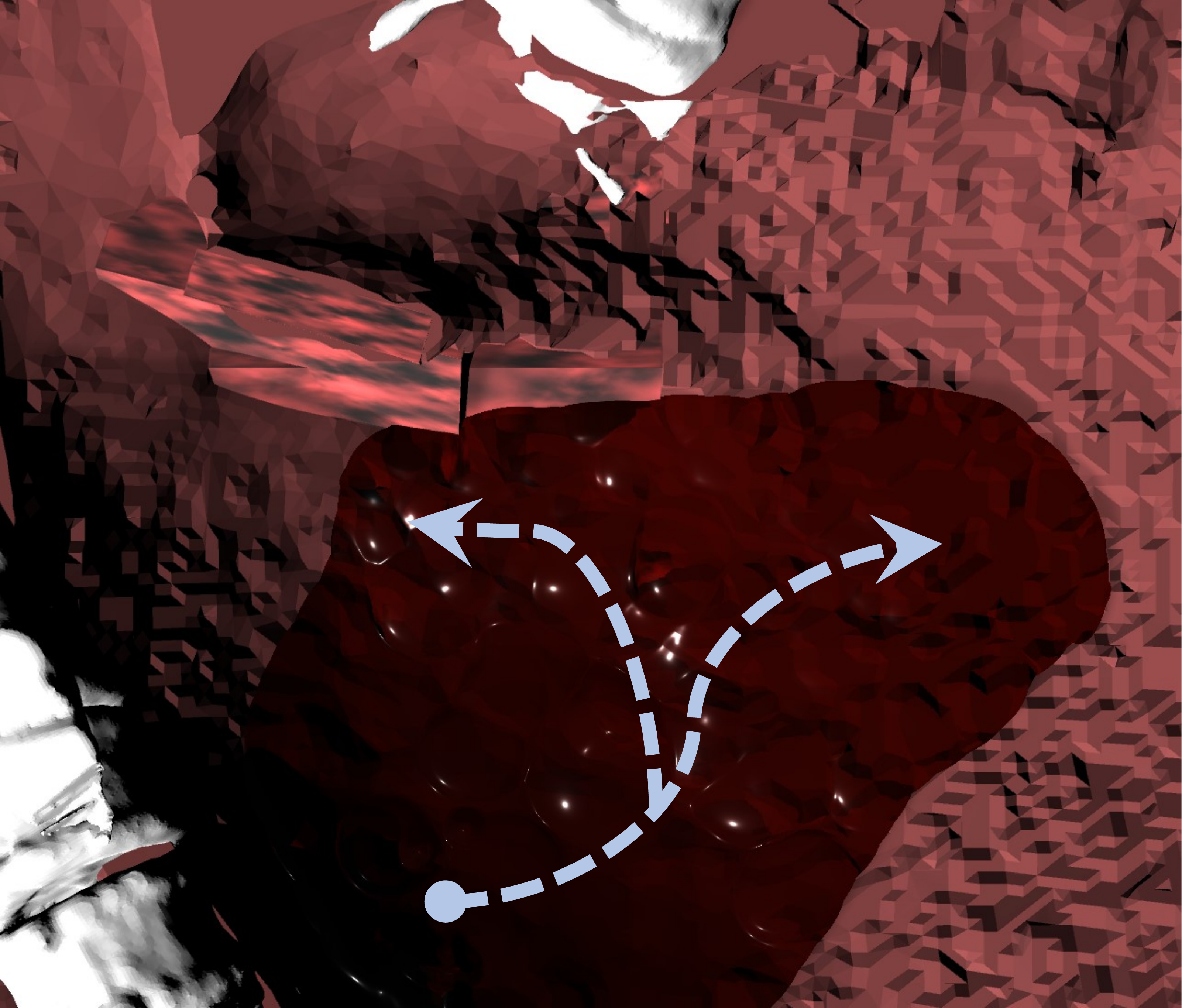}
	\end{subfigure} 
    \caption{Simulated scenes used to evaluate the proposed blood flow detection algorithm. The arrows highlight the direction of flow for the blood and how it fills the cavity. The detection algorithm estimates the blood flow via temporal tracking and aids in trajectory generation for a suction tool to remove the blood.}
    \label{fig:simulated_scenes}
    % \vspace{-2mm}
\end{figure*}

\begin{figure*}[t]
    % \vspace{2mm}
	\includegraphics[trim=2.5cm 1.5cm 3cm 1cm, clip, width=0.495\textwidth]{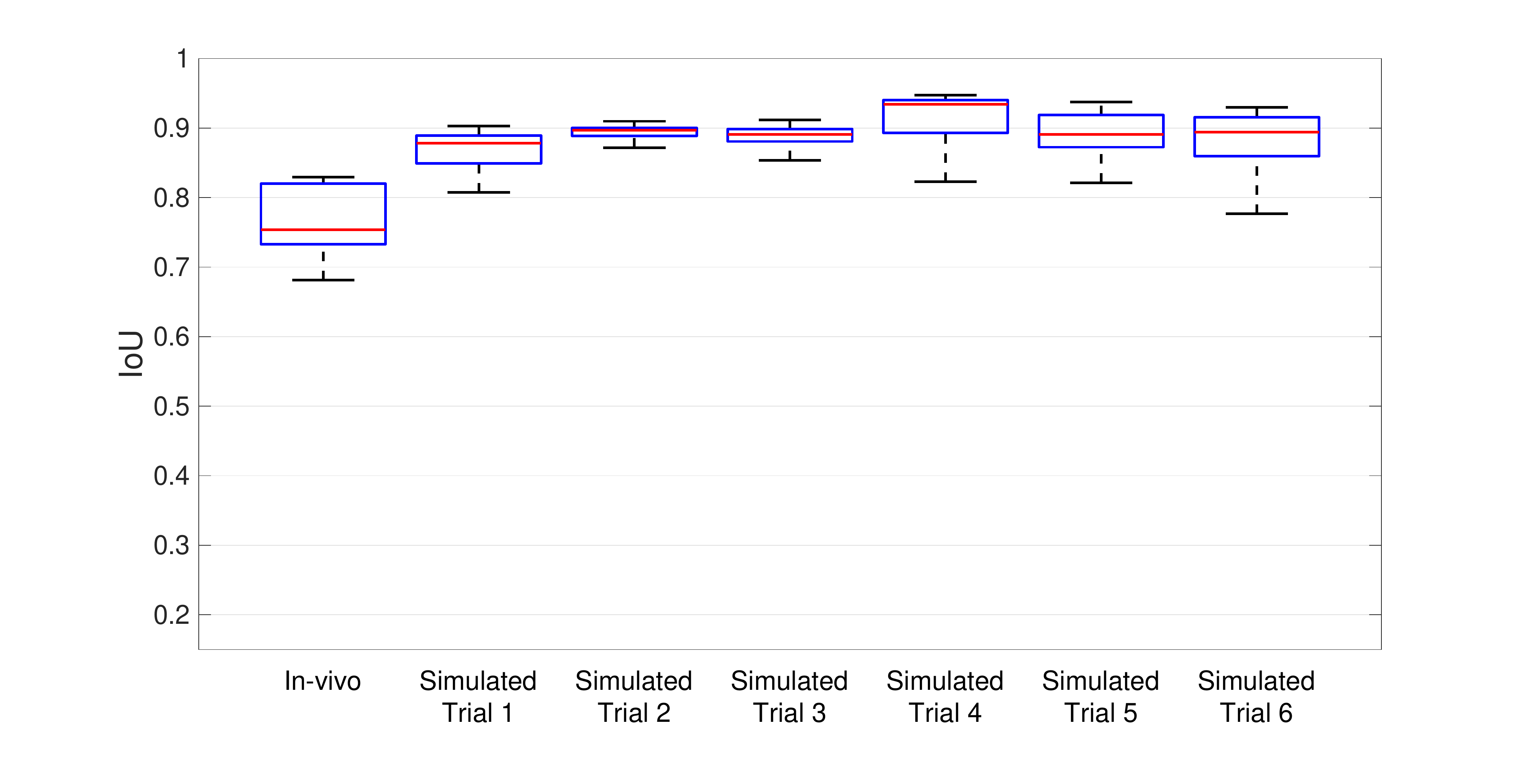}
	\includegraphics[trim=2.5cm 1.5cm 3cm 1cm, clip, width=0.495\textwidth]{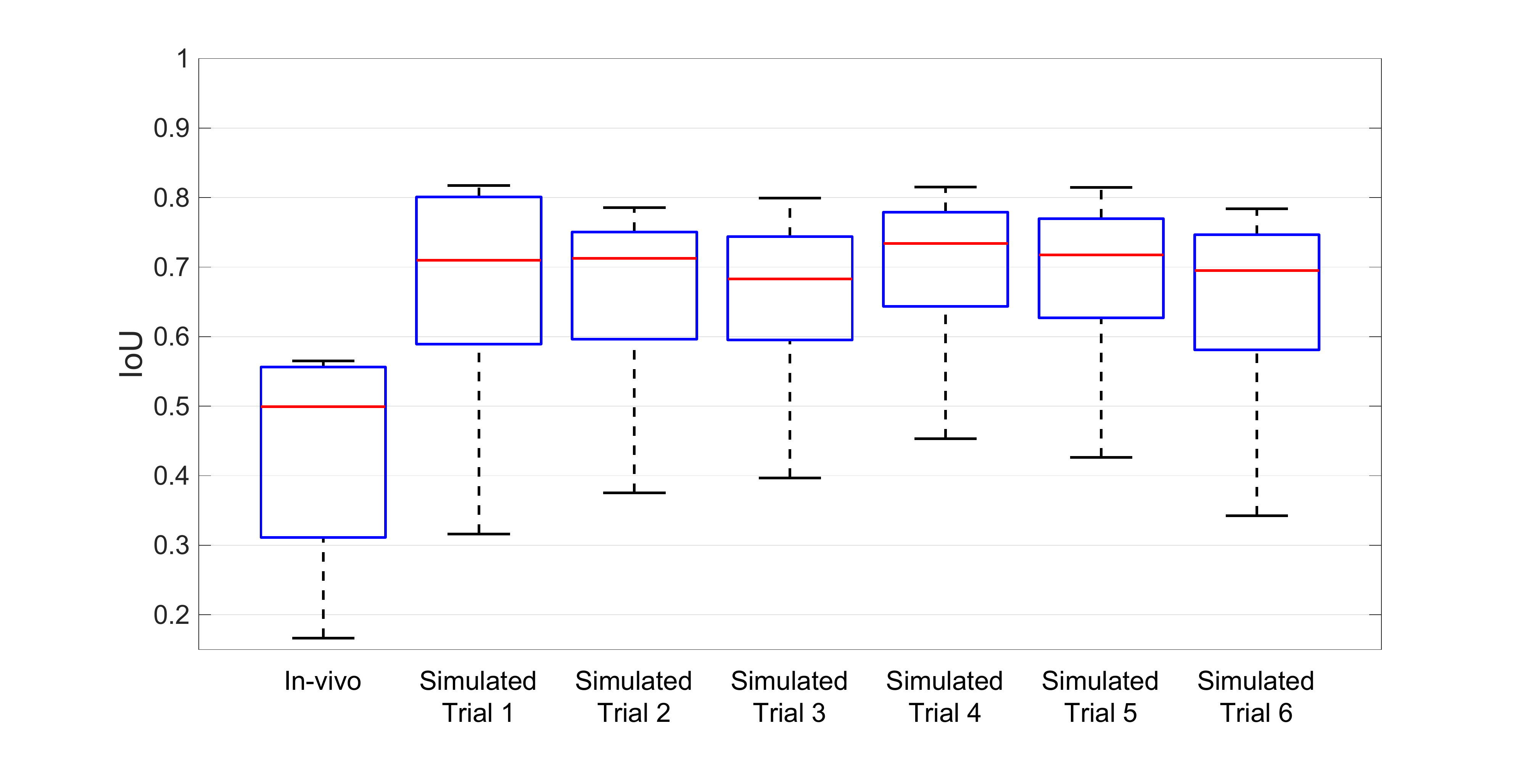}
	\caption{The left and right plots show Intersection over Union (IoU) results of blood flow estimation when using the proposed blood flow detection and tracking algorithm and only the blood detection from optical flow thresholding respectively. The complete pipeline results in better performance overall.}
    \label{fig:iou_results}
\end{figure*}

\subsection{Trajectory Generation for Blood Suction}

A start and end point must be decided to generate a trajectory for suction.
The end point should be roughly near the location of the vessel rupture in order to continuously remove any newly released blood.
Meanwhile, the starting point should be downstream of the flowing blood in order to effectively clear the surgical field when suctioning upstream towards the source.
Therefore, a simple estimation for the start and end point is done based on the age of the pixels in the blood region.
The pixel with the largest and smallest ages in the blood region are defined as the end and start points respectively.
To ensure that the end point is not generated at the exact edge of the blood stream, the blood region is eroded before selecting it.%, thereby \textit{skeletonizing} the detected blood regions so that tools that follow the skeletonized path will be more efficient at sucking up pools of blood with minimal instrument motion.

The trajectory generated from the start to end point should also maximize its ability to suction blood while moving upstream.
Therefore, using standard minimum distance paths are not ideal as they would tend to plan towards the edges of the blood region rather than the center.
To center the trajectory in the blood region, an additional clearance reward is given to the motion planner.
The clearance reward is generated by iteratively eroding the blood region for a max of $\gamma_r$ iterations.
The pixels left in the eroded region are given an additional reward of $r$ at each iteration.
The final trajectory in the image frame is then generated using Dijkstra's algorithm where the path is constrained to stay within the blood region and the clearance reward is subtracted from the normal distance cost.
An outline of this trajectory generation technique is shown in Algorithm \ref{alg:traj}, and an example is shown in Fig. \ref{fig:trajectory_gen_example}.
The trajectory is then executed if it is longer than a threshold $\gamma_T$.
This threshold gives time for the start and end points to stabilize so an effective trajectory can be generated.

\section{Experiments and Results}

The proposed blood flow detection and tracking method was evaluated on both simulated scenes and a live surgery involving a hemorrhage during a thyroidectomy.
The complete automated suctioning solution was demonstrated on in a lab setting on a simulated surgical cavity and red fluid for blood. The following sections describe these experiments, the necessary implementation details, and results.

\subsection{Implementation Details}
All subsequent experiments were ran on a computer with Intel\textregistered{} Core$^{\text{TM}}$ i9-7940X Processor and NVIDIA’s GeForce RTX 2080.
The blood flow detection and trajectory generation algorithms were implemented in MATLAB.
The CNN for optical flow estimation \cite{teney2016learning} is pre-trained on the Middlebury dataset \cite{baker2011database}, uses $l=3$ image frames for input,
and the resolution of the optical flow estimation is 1/4 of the input frame resolution.
These are the default values of the original CNN implementation.
The size of the probability map is set to the optical flow resolution.
The threshold for detection, $\gamma_o$, region size, $\gamma_B$, maximum number of erosions for clearance, $\gamma_r$, and trajectory length, $\gamma_T$, are set to 0.45, 20, 4, and 30 respectively.
The detection probability, $P(z^p_t = \text{b} | p_t = \text{b})$, $P(z^p_t = \text{b} | p_t = \overline{\text{b}})$, are set to 0.95 and 0.2 respectively because experimentally we found the true positive rate and false negative rate to be very accurate and noisy respectively.
The initial probability of a pixel being blood, $P(p_0 = \text{b})$ and transition probabilities of a pixel being blood, $P(p_{t+1} = \text{b} | p_t = \text{b})$, $P(p_{t+1} = \text{b} | p_t = \overline{\text{b}}, k^p_t = \text{b})$, $P(p_{t+1} = \text{b} | p_t = \overline{\text{b}}, k^p_t = \overline{\text{b}})$, are set to 0.1, 0.98, 0.85, and 0.01 respectively.
The neighbors for a pixel, $\mathbf{A}_p$, are set to just up, down, left, and right since the algorithm needs to run quickly for real-time detection in the upcoming experiments.
The clearance reward per erosion, $r$, is set to 0.2.

\begin{figure*}[t]
    \centering
    \vspace{2mm}
	\begin{subfigure}{0.245\textwidth}
	\includegraphics[width=1\textwidth]{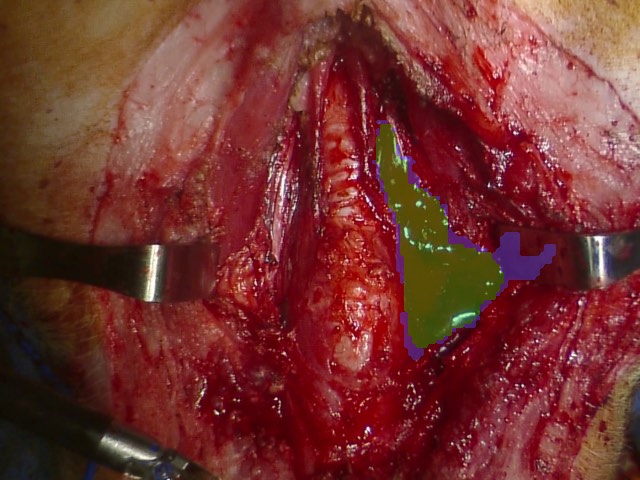}
	\end{subfigure}
	\begin{subfigure}{0.245\textwidth}
	\includegraphics[width=1\textwidth]{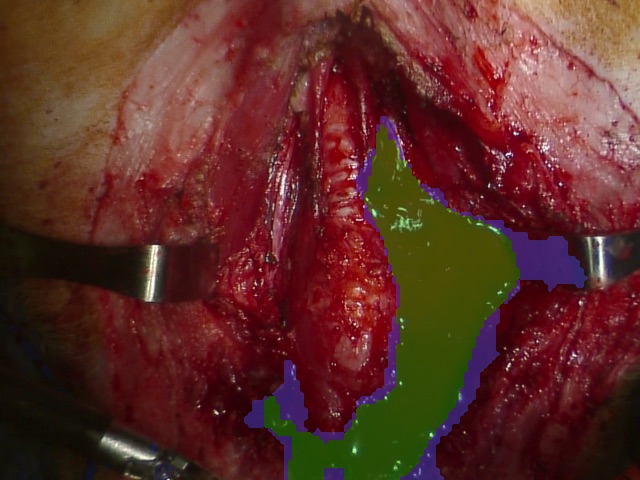}
	\end{subfigure} 
	\begin{subfigure}{0.245\textwidth}
	\includegraphics[width=1\textwidth]{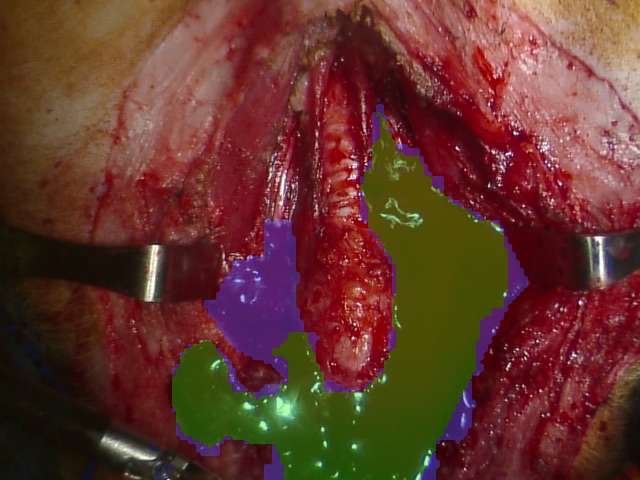}
	\end{subfigure} 
	\begin{subfigure}{0.245\textwidth}
	\includegraphics[width=1\textwidth]{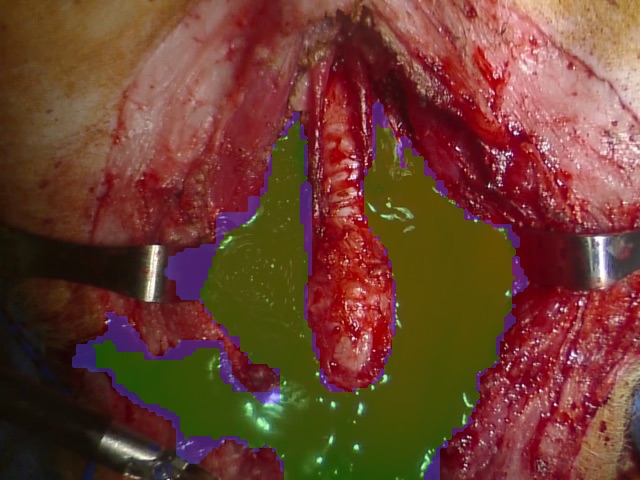}
	\end{subfigure}
	
	\vspace{1mm}
	
	\begin{subfigure}{0.245\textwidth}
	\includegraphics[width=1\textwidth]{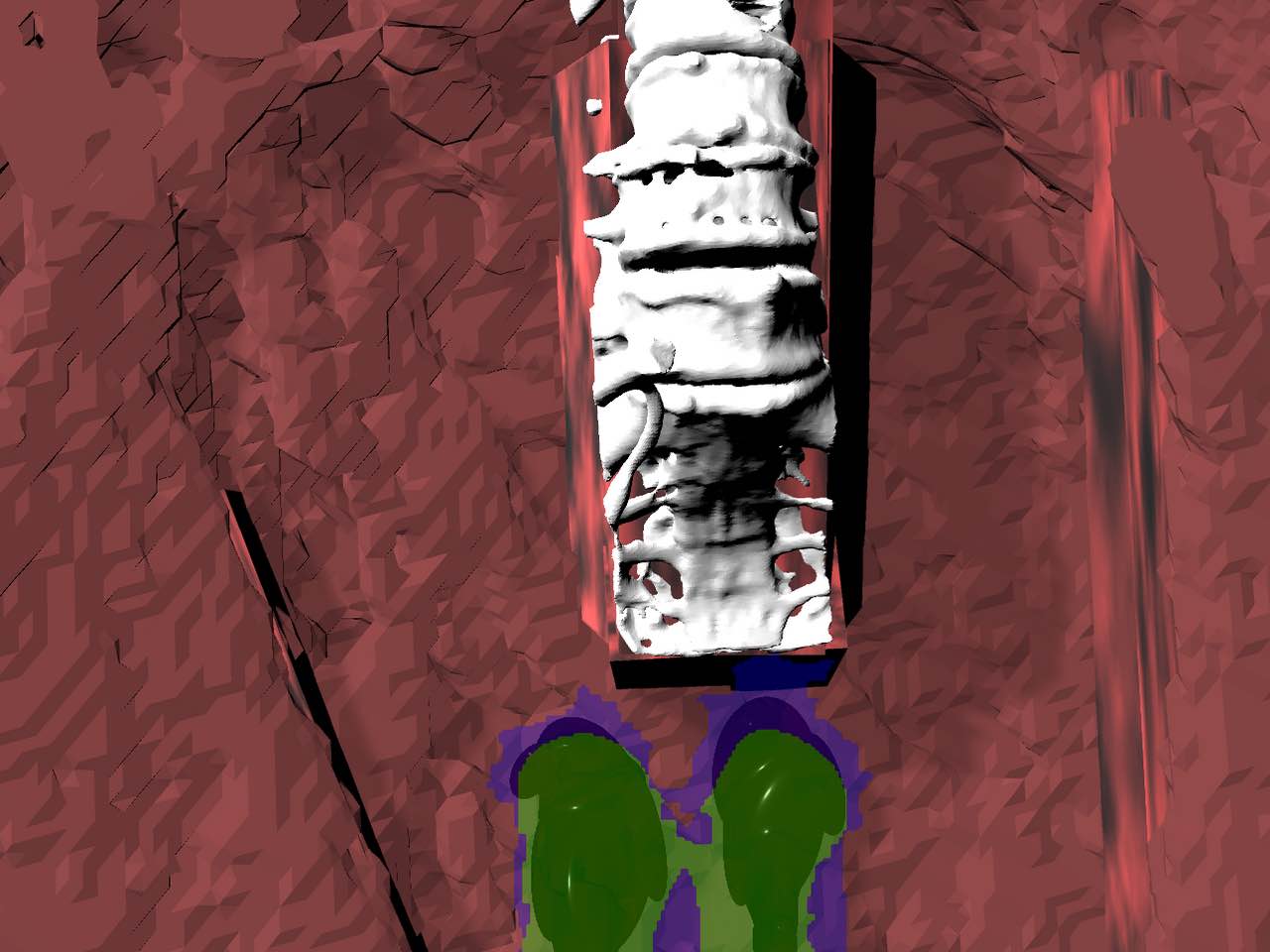}
	\end{subfigure}
	\begin{subfigure}{0.245\textwidth}
	\includegraphics[width=1\textwidth]{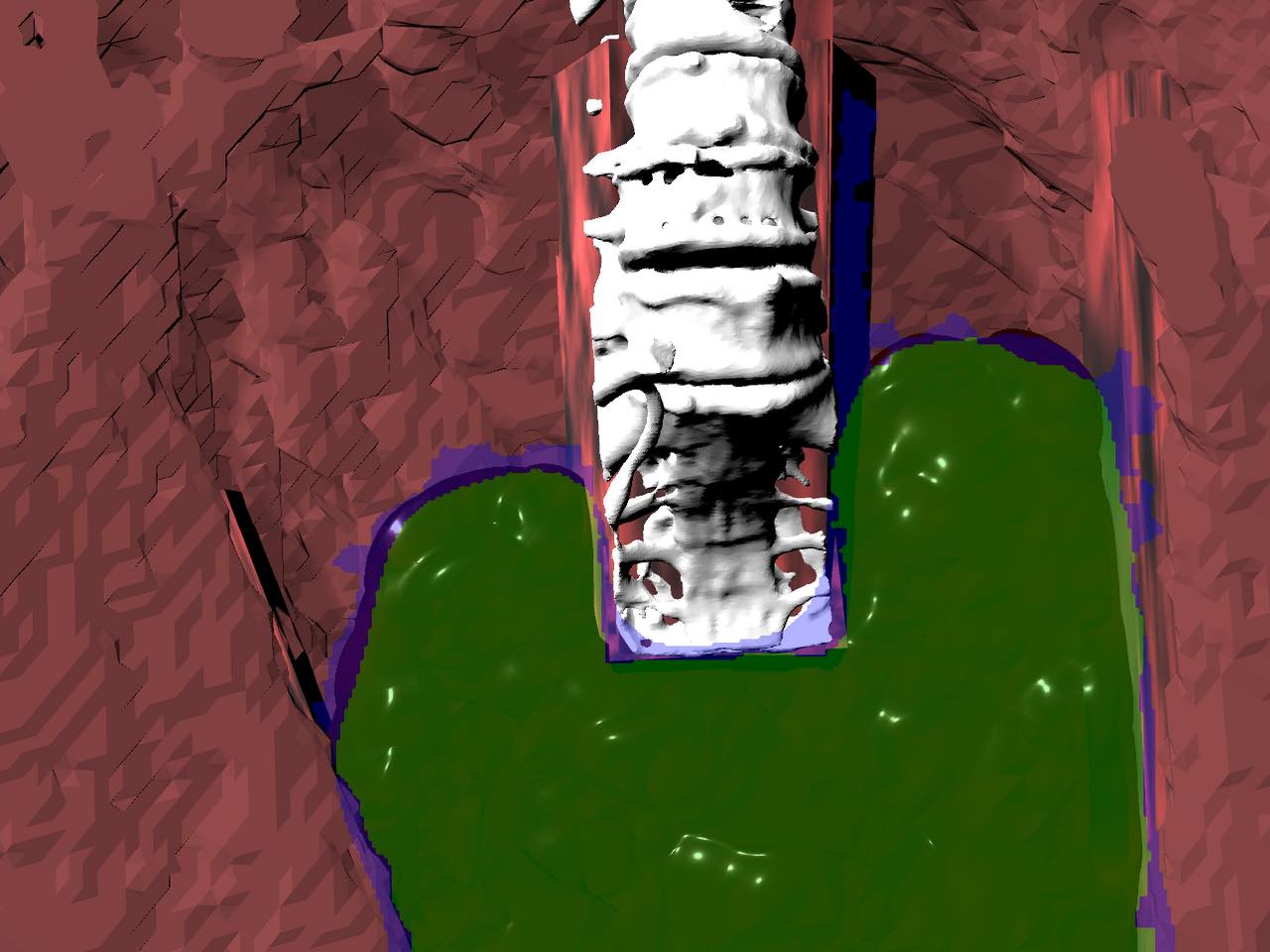}
	\end{subfigure} 
	\begin{subfigure}{0.245\textwidth}
	\includegraphics[width=1\textwidth]{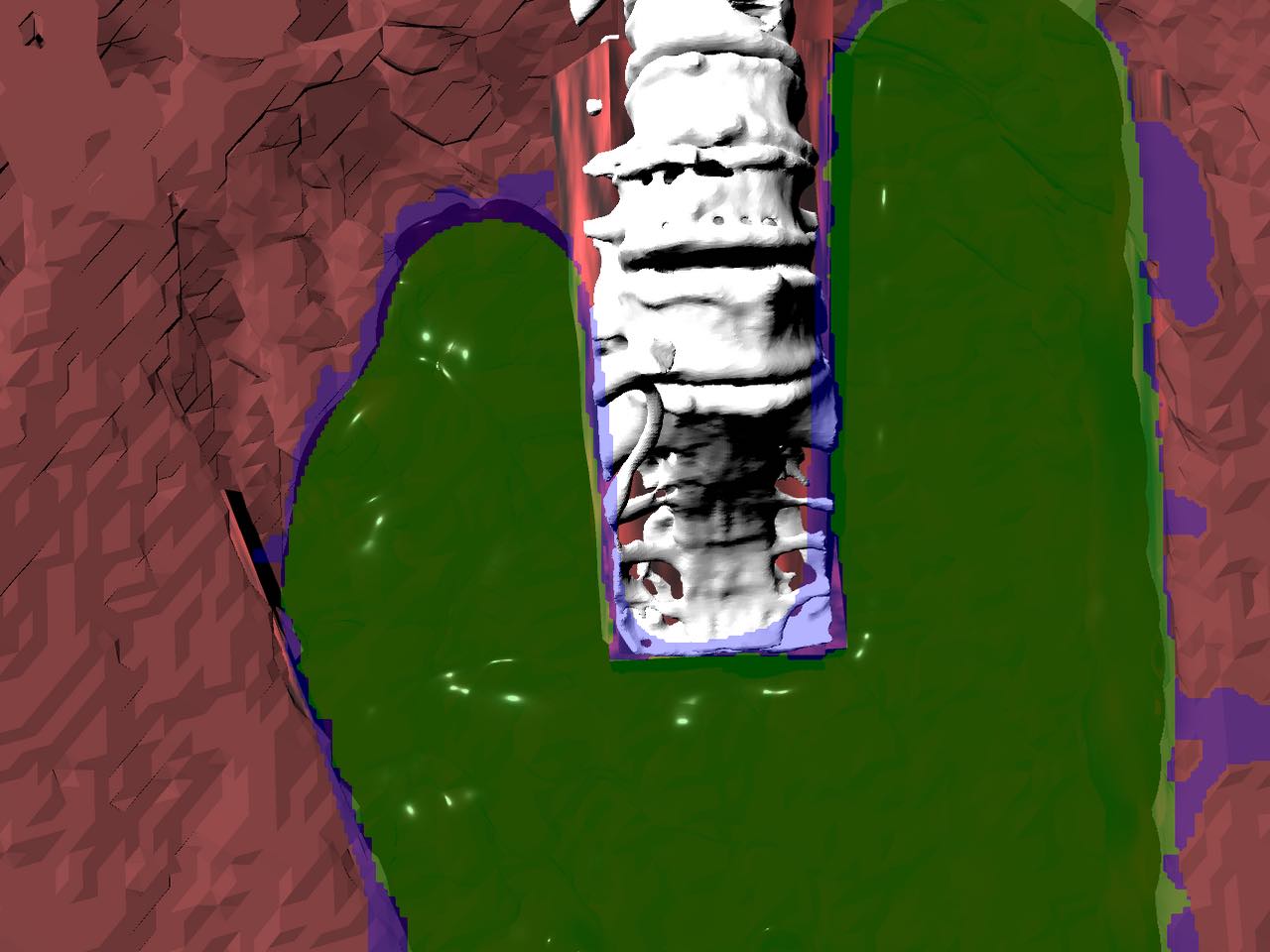}
	\end{subfigure} 
	\begin{subfigure}{0.245\textwidth}
	\includegraphics[width=1\textwidth]{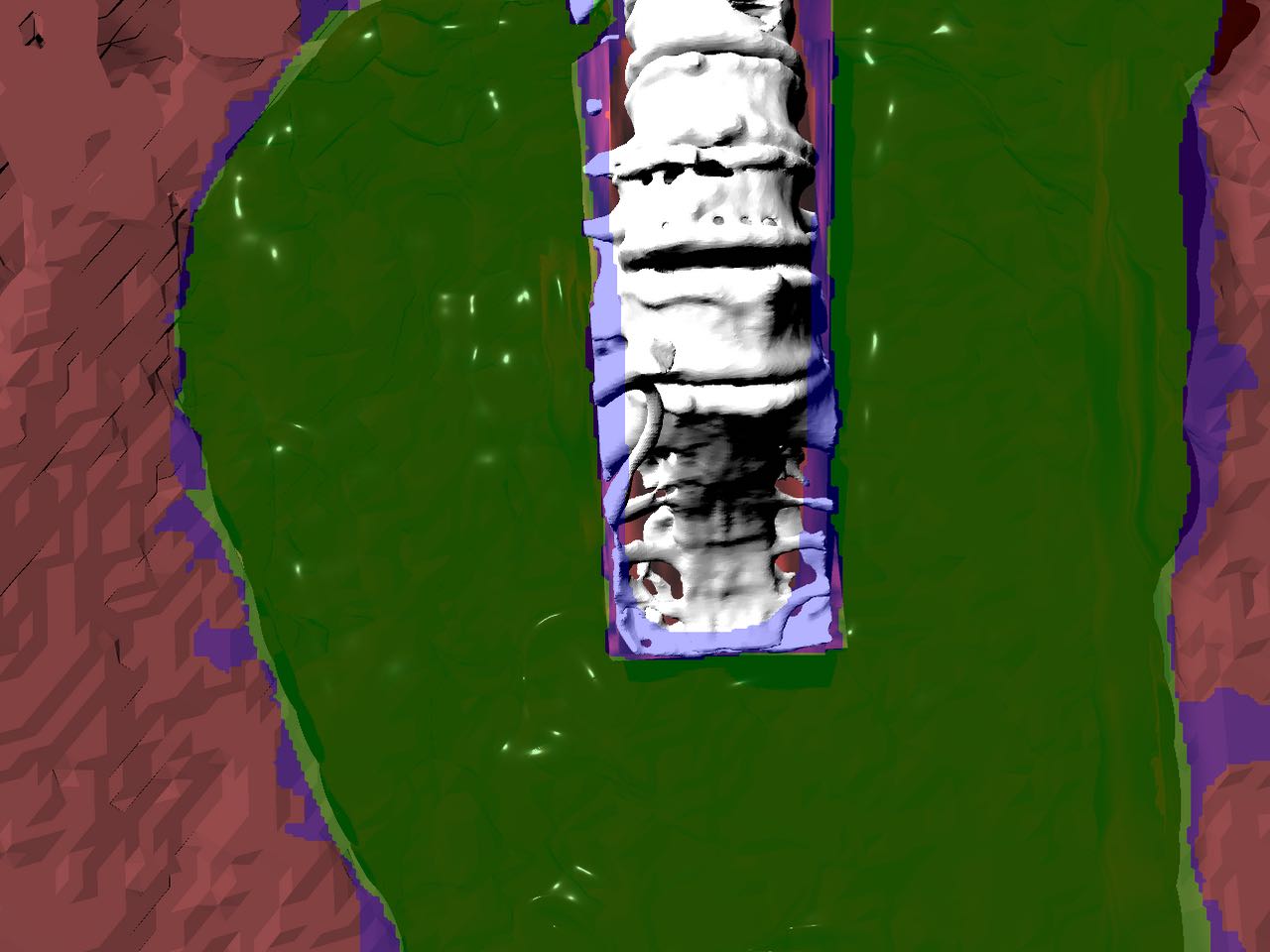}
	\end{subfigure}
    \caption{Example figures of blood flow detection and tracking. The top and bottom sequences are from an in-vivo rupture and simulated scene respectively. The highlighted green regions are the intersection between ground truth and blood flow detection. Meanwhile, the highlighted blue regions are the error of detection, i.e., the union minus intersection between the ground truth and blood flow detection. These figures are best viewed in color.% Note that utilizing temporal information allows for consistent detection of blood flow even when there are previous red blood stains in the surgical field.
    }
    \label{fig:IoU_qualitive_results}
\end{figure*}

\subsection{Datasets to Evaluate Blood Region Detection}
Two separate datasets were generated for this work to evaluate the proposed blood region detection algorithm.
Both datasets have labelled ground-truth masks, $\mathbf{G}_t$, of the blood region.
Performance is evaluated from these datasets using the Intersection over Union (IoU) metric:
\begin{equation}
    \frac{\mathbf{B}_t \wedge \mathbf{G}_t}{\mathbf{B}_t \vee \mathbf{G}_t}
    \label{eq:iou}
\end{equation}
where $\mathbf{B}_t$ is a mask of the detected blood region from our proposed method, $\wedge$ is the Boolean-AND operation and $\vee$ is the Boolean-OR operation.

\subsubsection{Simulated Scenes}
Six simulated scenes of flowing blood are generated using Unity3D's particle-based fluid dynamics (PBDs).
The scenes are shown in Fig. \ref{fig:simulated_scenes}. A total of 61 image frames were extracted per scene.
The ground-truth mask, $\mathbf{G}_t$, of the blood region is generated by projecting the fluid particles onto a virtual camera's image plane and applying Gaussian smoothing.
% Performance is evaluated using Intersection over Union (IoU) as a metric.
% The IoU is calculated as:
% \begin{equation}
%     \frac{\mathbf{B}_t \wedge \mathbf{G}_t}{\mathbf{B}_t \vee \mathbf{G}_t}
%     \label{eq:iou}
% \end{equation}
% where $\mathbf{B}_t$ is a mask of the detected blood region from our proposed method, $\wedge$ is the Boolean-AND operation and $\vee$ is the Boolean-OR operation.
The rendered image is set to 1095$\times$1284 pixels.

\subsubsection{In-Vivo Surgical Scene}
After the completion of a thyroidectomy conducted on a pig (UCSD IACUC S19130), a rupture occurred on the carotid artery.
The 8s video data from this incident is used to evaluate the blood flow detection and tracking algorithm in a similar manner to the simulated scenes.
For ground-truth masks of the blood region, $\mathbf{G}_t$, 10 evenly distributed frames were manually annotated.
The recorded image size was 640 by 480 pixels.

\subsection{Performance of Blood Region Detection}

To show the effectiveness of the tracking algorithm, a comparison experiment was conducted where the blood flow region was simply set to be the pixels with optical flow magnitude greater than $\gamma_O$.
The distribution of IoU results are shown in Fig. \ref{fig:iou_results} for blood flow region detection with and without the tracking algorithm on the simulated scenes and in-vivo dataset.
There is a clear difference in performance of the blood flow detection and tracking between the simulated scenes and in-vivo rupture.
We believe this is due to the poorer lighting conditions and the reflective surgical clamps used in the in-vivo scene as seen in Fig. \ref{fig:IoU_qualitive_results}.
Nonetheless, the blood region is successfully estimated when using the tracking algorithm despite the many red stains, hence highlighting the importance of using temporal information for detection rather than color features.
For additional comparison, Lucas-Kanade's \cite{lucas1981iterative} and Farnebeck's \cite{farneback2003two} optical flow estimation techniques were replaced for the CNN based optical flow estimation \cite{teney2016learning}.
Note that Lucas-Kanade's optical flow estimation is the proposed detection method for fluids by Yamaguchi and Atkeson used for robot pouring \cite{yamaguchi2016stereo}.
The resulting IoU for the in-vivo and simulated scenes was measured to be consistently below 0.50 in both cases, which is substantially lower than our proposed detection technique.

\begin{figure}[t]
    \centering
	\includegraphics[trim=0cm 0cm 6.65cm 0cm, clip, width=0.45\textwidth]{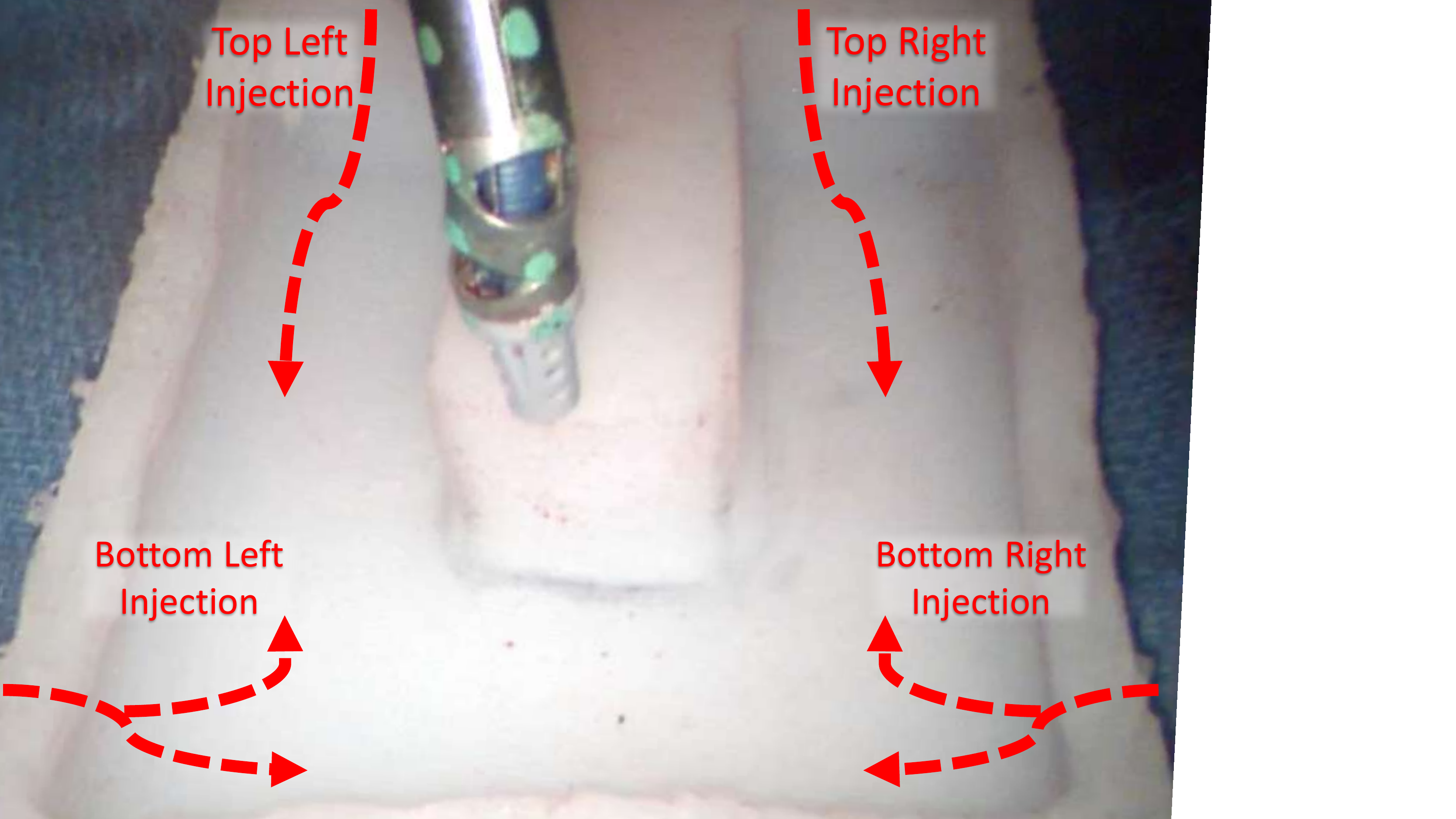}
	\caption{Endoscopic view of phantom used for the automated suction experiments. The red arrows highlight the direction of flow at the four injection points tested in the experiment.}
	\vspace{-2mm}
    \label{fig:silicon_cavity}
\end{figure}

\begin{figure*}[t]
    \centering
    \vspace{2mm}
    \begin{subfigure}{0.24\textwidth}
	\includegraphics[width=1\textwidth]{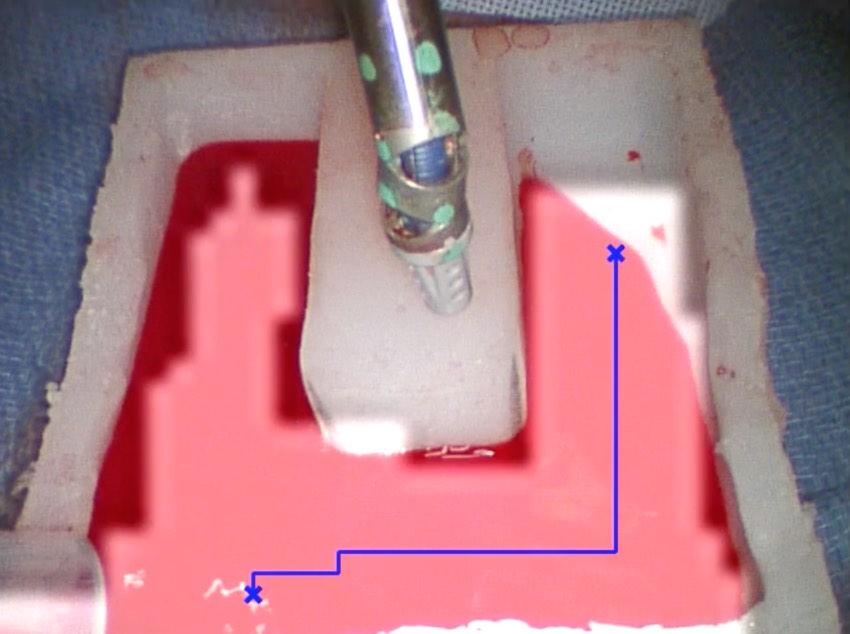}
	\end{subfigure}
	\begin{subfigure}{0.24\textwidth}
	\includegraphics[width=1\textwidth]{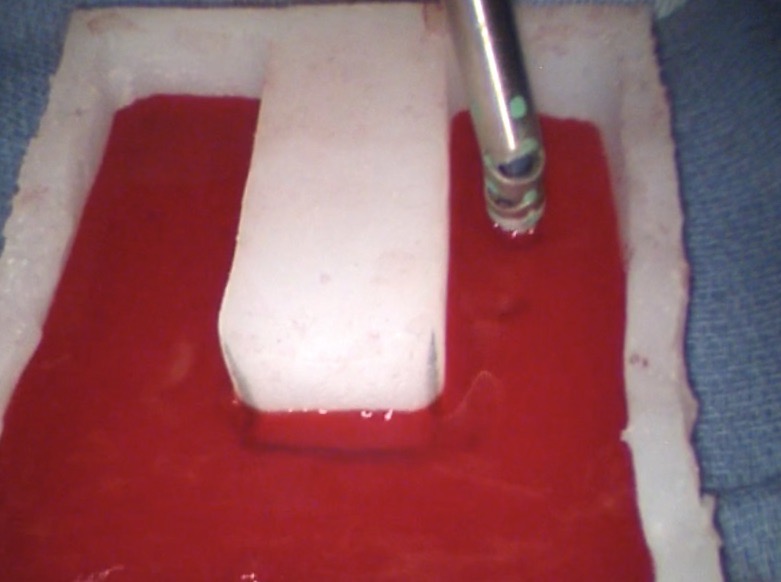}
	\end{subfigure}
	\begin{subfigure}{0.24\textwidth}
	\includegraphics[width=1\textwidth]{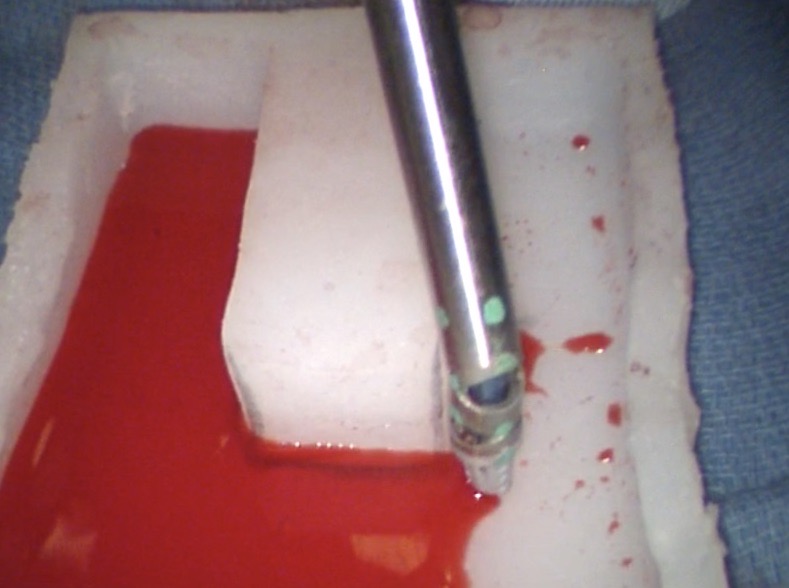}
	\end{subfigure}
	\begin{subfigure}{0.24\textwidth}
	\includegraphics[width=1\textwidth]{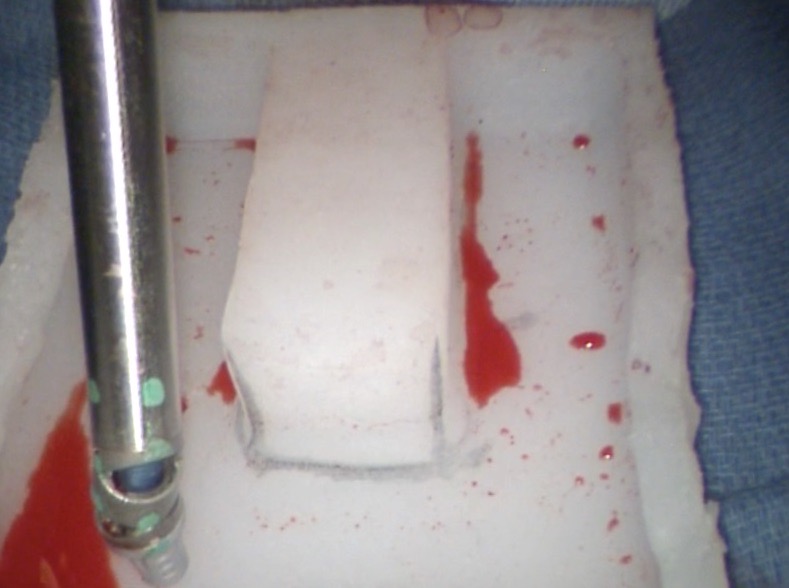}
	\end{subfigure}
    
%     \vspace{1mm}
    
%     \begin{subfigure}{0.24\textwidth}
% 	\includegraphics[width=1\textwidth]{Figures/simple_cavity/sequence_0.jpg}
% 	\end{subfigure}
% %     \begin{subfigure}{0.19\textwidth}
% % 	\includegraphics[width=1\textwidth]{Figures/simple_cavity/sequence_1.jpg}
% % 	\end{subfigure}
% 	\begin{subfigure}{0.24\textwidth}
% 	\includegraphics[width=1\textwidth]{Figures/simple_cavity/sequence_2.jpg}
% 	\end{subfigure}
% 	\begin{subfigure}{0.24\textwidth}
% 	\includegraphics[width=1\textwidth]{Figures/simple_cavity/sequence_3.jpg}
% 	\end{subfigure} 
% 	\begin{subfigure}{0.24\textwidth}
% 	\includegraphics[width=1\textwidth]{Figures/simple_cavity/sequence_4.jpg}
% 	\end{subfigure} 
    \caption{Sequence of figures (from left to right) of an automated suction experiment where the liquid is injected at the bottom left corner.
    The images are captured from the endoscopic view with the first one highlighting the detected and tracked blood region in white and the generated trajectory in blue.
    To achieve real-time capabilities and react quickly to flowing blood, the blood detection and tracking algorithm is set to a lower resolution in these experiments.
    Nonetheless, the robotic suction tool follows the generated trajectory and efficiently removes fluids from the cavity.
    % The top row shows the endoscopic view with the first one highlighting the detected and tracked blood region in white and the generated trajectory in blue.
    }
    \label{fig:simple_cavity_suction}
\end{figure*}

\subsection{Automated Suction in Cavity}

To evaluate the complete autonomous surgical task of recognizing blood flow and performing autonomous suction, a tissue phantom with a cavity for liquid to flow through is constructed from silicone, and water with red coloring dye is drained into the cavity using surgical tubing as shown in Fig. \ref{fig:silicon_cavity}.
A stereoscopic camera with a resolution of 1080$\times$1920 pixels at 30fps on the dVRK \cite{kazanzides2014open} is used.
The trajectory generated for suction is converted into 3D position commands using the Pyramid Stereo Matching Network (PSMNet) \cite{chang2018pyramid}, which takes the stereo-images of the cameras and determines the depth of each pixel.
PSMNet's weight are provided by their original implementations without any task-specific fine-tuning; the maximum disparity is set to 192.
PSMNet estimated depth using an image size of 640 by 480 pixels meanwhile the blood flow detection algorithm used a reduced image size of 160 by 120 pixels to improve its speed.

A Patient Side Manipulator (PSM) from the dVRK \cite{kazanzides2014open} was equipped with da Vinci\textregistered{} Si Suction Tool and followed the generated trajectory to clear the simulated surgical cavity from blood.
To follow the trajectory, the position of the end-effector in the PSM base frame, $\mathbf{b}_t$, is iteratively set to:
\begin{equation}
    \overline{\mathbf{b}}_{t+1} = \begin{cases} \gamma_s \frac{\mathbf{d}_{t}}{||\mathbf{d}_{t}||} + \overline{\mathbf{b}}_{t}   & \text{if } ||\mathbf{d}_{t}|| > \gamma_s\\
    \mathbf{d}_{t} + \overline{\mathbf{b}}_{t} & \text{if } ||\mathbf{d}_{t}|| \leq \gamma_s
    \end{cases}
\end{equation}
where $\gamma_s = 0.75$mm is the max step size, the operator $\overline{\cdot} = \begin{bmatrix} \cdot& 1\end{bmatrix}^\top$ gives the homogeneous representation of a point, and the direction, $\mathbf{d}_t$, is computed as
\begin{equation}
    \mathbf{d}_t = \mathbf{T}^b_c \overline{\mathbf{b}}^g - \overline{\mathbf{b}}_{t}
\end{equation}
The camera to base transform, $\mathbf{T}^b_c \in SE(3)$, is estimated in real time using our previous work \cite{li2020super} and $\mathbf{b}^g$ is the 3D goal position generated by the trajectory and PSMNet.
This controller is ran at 100Hz and is repeated until $||\mathbf{d}_t|| < 2$mm per target position, $\mathbf{b}^g$, from the generated trajectory.
Meanwhile the orientation of the suction tip is set to always be in direction of gravity.
The position, $\mathbf{b}_t$, and orientation of the end-effector is converted to joint angles using an analytical inverse kinematic solution.
These joint angles are then regulated using dVRK \cite{kazanzides2014open}.

To account for imperfections in the 3D depth estimation from PSMNet and surgical tool tracking to regulate the end-effector, the suction tool was commanded to oscillate in and out along the direction of gravity an additional 5mm at every point on the trajectory.
This probing behavior ensured that the tool always sucked up the blood and neither drifted above the blood nor penetrated and dragged tissue.
The Robot Operation System (ROS) is used to encapsulate the image processing and robot trajectory tracking processes \cite{ros}.

Roughly 50mL of liquid is injected using a syringe into the cavity at one of the four locations highlighted in Fig. \ref{fig:silicon_cavity}.
Before each trial, the end-effector is centered in the middle of the silicon mold such that it does not obstruct any of the injected liquid as shown in Fig. \ref{fig:silicon_cavity}.
The percentage of liquid removed from the cavity, time to react to the injected fluid, the  time to complete the trajectory were measured to evaluate the performance of the proposed automation method.
The percentage of liquid removed was measured by weighing the silicon mold and syringe before and after each trial.
Time to react refers to the time taken to detect the flowing blood and generate a trajectory (i.e., completing Algorithm \ref{alg}) from the first moment that the injected blood is visible in the camera frame.
To ensure consistency of the proposed automation method, the experiment is repeated ten times at each of the four injection spots.

The results from the total 40 trials of the automated suction experiment are shown in Table \ref{tab:automated_suction_results} and an example sequence is shown in Fig. \ref{fig:simple_cavity_suction}.
During the experiments, we noticed the liquid generally pooled near the bottom left injection point since it was slightly lower with respect to gravity than the rest of the cavity.
This lead to shorter trajectories being generated, and hence less time to execute them as seen in the results, for the bottom left corner experiment compared to the others.

A similar set up is repeated for demonstration purposes and the result is shown in Fig. \ref{fig:cover_figure}.
The mold used for this demonstration was constructed using candle wax and has pig intestine embedded in it.
Everything else is kept the same as the previously described.
Despite changes to the visual textures and topology of the scene, we can see that the method is robust enough to perform autonomous tracking and suction of blood without modification.

\begin{table}[t]
    \centering
    \caption{Mean results from automated suction experiment at each injection point. }
    \def\arraystretch{1.2}
    \begin{tabular}{c|c|c|c}
      \multirow{ 2}{*}{\textbf{Injection Point}}  & \textbf{Liquid} & \multirow{ 2}{*}{\textbf{Time to React}}  & \textbf{Time to Execute}\\
         & \textbf{Suctioned} & & \textbf{Trajectory} \\ \hline
         Top Left & 96.7\% & 4.1s & 45.3s \\
         Top Right & 93.6\% & 2.6s & 47.4s \\
         Bottom Left & 96.1\% & 4.7s & 23.9s \\
         Bottom Right & 96.8\% & 6.4s & 38.1s 
    \end{tabular}
    \label{tab:automated_suction_results}
\end{table}

\section{Discussion and Conclusion}

In this work, the first completely automated solution for clearing the surgical field from blood is presented. 
The solution provides both the perception, trajectory generation, and control strategy required for the task of clearing blood.
This is the first step taken towards automation of a crucial surgical task, hemostasis, which can occur in any surgery at any time.
To ensure robustness against blood stains, the algorithm relies on temporal information for detection.
The novel blood flow detection and tracking algorithm presented offers a unique, probabilistic solution to tracking liquids over 3D cavities and channels, under noisy and harsh visibility conditions, and is a critical perceptual element.
This estimation and tracking helps inform a trajectory generation technique to act upon the detected blood and uses a clearance reward to maximize the blood suctioned by the suction tool and be robust against imperfect blood region estimation.

For future work, we intend to push towards a complete solution for automation of hemostasis.
To accomplish this, a more precise location of the bleeding source will be estimated using a particle based motion model, similar to PBD simulators, in the temporal filtering.
Another consideration will be integration of surgical tool masking using our previous works \cite{li2020super, lu2020super} into the blood tracking framework as the hemostasis automation task will require additional surgical robotic tools.
% Finally, we will investigate automation techniques using the additional surgical robotic tools to close the identified vessel rupture.

% For future work, we intend to push towards a complete solution for automation of hemostasis.
% To accomplish this, a more precise estimation of the source is needed which will be investigated by using a particle based motion model, similar to particle based physics simulators, in the temporal filtering.
% This should improve the blood flow tracking performance since it will have a more exact prediction step.
% In addition, the motion model can also be used as a forward prediction for model predictive control techniques so the suction tool can anticipate, rather than reacting, to the blood flow.
% Fine tissue manipulation policies that have not been explored, such as vessel grasping and clipping, will be investigated, likely leveraging open-sourced platforms like the da Vinci Reinforcement Learning (dVRL) toolkit \cite{richter2019open}.

\section{Acknowledgements}
This work is supported by the National Science Foundation (NSF) under grant number 1830403 and 1935329 and the US Army Telemedicine and Advanced Technology Research Center (TATRC) under the Robotic Battlefield Medical Support System project.
F. Richter is supported by the NSF Graduate Research Fellowship.
The authors would like to thank Intuitive Surgical Inc. for instrument donations, Jingpei Lu for his support with PSMNet, Harleen Singh for her support with the molds, Simon DiMiao, Omid Maherari, Dale Bergman, and Anton Deguet for their support with the dVRK.

\balance
\bibliographystyle{ieeetr}
\bibliography{references}

\begin{thebibliography}{10}

\bibitem{yipDasJournal}
M.~Yip and N.~Das, ``Robot autonomy for surgery,'' in {\em Encyclopedia of
  Medical Robotics}, ch.~10, pp.~281--313, World Scientific, 2017.

\bibitem{kazanzides2014open}
P.~Kazanzides, Z.~Chen, A.~Deguet, G.~S. Fischer, R.~H. Taylor, and S.~P.
  DiMaio, ``An open-source research kit for the da vinci{\textregistered}
  surgical system,'' in {\em 2014 IEEE international conference on robotics and
  automation (ICRA)}, pp.~6434--6439, IEEE, 2014.

\bibitem{richter2019open}
F.~Richter, R.~K. Orosco, and M.~C. Yip, ``Open-sourced reinforcement learning
  environments for surgical robotics,'' {\em arXiv preprint arXiv:1903.02090},
  2019.

\bibitem{murali2015learning}
A.~Murali, S.~Sen, B.~Kehoe, A.~Garg, S.~McFarland, S.~Patil, W.~D. Boyd,
  S.~Lim, P.~Abbeel, and K.~Goldberg, ``Learning by observation for surgical
  subtasks: Multilateral cutting of 3d viscoelastic and 2d orthotropic tissue
  phantoms,'' in {\em 2015 IEEE International Conference on Robotics and
  Automation (ICRA)}, pp.~1202--1209, IEEE, 2015.

\bibitem{thananjeyan2017multilateral}
B.~Thananjeyan, A.~Garg, S.~Krishnan, C.~Chen, L.~Miller, and K.~Goldberg,
  ``Multilateral surgical pattern cutting in 2d orthotropic gauze with deep
  reinforcement learning policies for tensioning,'' in {\em 2017 IEEE
  International Conference on Robotics and Automation (ICRA)}, pp.~2371--2378,
  IEEE, 2017.

\bibitem{d2018automated}
C.~D'Ettorre, G.~Dwyer, X.~Du, F.~Chadebecq, F.~Vasconcelos, E.~De~Momi, and
  D.~Stoyanov, ``Automated pick-up of suturing needles for robotic surgical
  assistance,'' in {\em 2018 IEEE International Conference on Robotics and
  Automation (ICRA)}, pp.~1370--1377, IEEE, 2018.

\bibitem{sen2016automating}
S.~Sen, A.~Garg, D.~V. Gealy, S.~McKinley, Y.~Jen, and K.~Goldberg,
  ``Automating multi-throw multilateral surgical suturing with a mechanical
  needle guide and sequential convex optimization,'' in {\em 2016 IEEE
  International Conference on Robotics and Automation (ICRA)}, pp.~4178--4185,
  IEEE, 2016.

\bibitem{zhong2019dual}
F.~Zhong, Y.~Wang, Z.~Wang, and Y.-H. Liu, ``Dual-arm robotic needle insertion
  with active tissue deformation for autonomous suturing,'' {\em IEEE Robotics
  and Automation Letters}, vol.~4, no.~3, pp.~2669--2676, 2019.

\bibitem{kehoe2014autonomous}
B.~Kehoe, G.~Kahn, J.~Mahler, J.~Kim, A.~Lee, A.~Lee, K.~Nakagawa, S.~Patil,
  W.~D. Boyd, P.~Abbeel, {\em et~al.}, ``Autonomous multilateral debridement
  with the raven surgical robot,'' in {\em 2014 IEEE International Conference
  on Robotics and Automation (ICRA)}, pp.~1432--1439, IEEE, 2014.

\bibitem{li2020super}
Y.~Li, F.~Richter, J.~Lu, E.~K. Funk, R.~K. Orosco, J.~Zhu, and M.~C. Yip,
  ``Super: A surgical perception framework for endoscopic tissue manipulation
  with surgical robotics,'' {\em IEEE Robotics and Automation Letters}, vol.~5,
  no.~2, pp.~2294--2301, 2020.

\bibitem{lu2020super}
J.~Lu, A.~Jayakumari, F.~Richter, Y.~Li, and M.~C. Yip, ``Super deep: A
  surgical perception framework for robotic tissue manipulation using deep
  learning for feature extraction,'' {\em arXiv preprint arXiv:2003.03472},
  2020.

\bibitem{liedlgruber2011computer}
M.~Liedlgruber and A.~Uhl, ``Computer-aided decision support systems for
  endoscopy in the gastrointestinal tract: a review,'' {\em IEEE reviews in
  biomedical engineering}, vol.~4, pp.~73--88, 2011.

\bibitem{fu2013computer}
Y.~Fu, W.~Zhang, M.~Mandal, and M.~Q.-H. Meng, ``Computer-aided bleeding
  detection in wce video,'' {\em IEEE journal of biomedical and health
  informatics}, vol.~18, no.~2, pp.~636--642, 2013.

\bibitem{liu2009obscure}
J.~Liu and X.~Yuan, ``Obscure bleeding detection in endoscopy images using
  support vector machines,'' {\em Optimization and engineering}, vol.~10,
  no.~2, pp.~289--299, 2009.

\bibitem{jung2008active}
Y.~S. Jung, Y.~H. Kim, D.~H. Lee, and J.~H. Kim, ``Active blood detection in a
  high resolution capsule endoscopy using color spectrum transformation,'' in
  {\em 2008 International Conference on BioMedical Engineering and
  Informatics}, vol.~1, pp.~859--862, IEEE, 2008.

\bibitem{okamoto2019real}
T.~Okamoto, T.~Ohnishi, H.~Kawahira, O.~Dergachyava, P.~Jannin, and
  H.~Haneishi, ``Real-time identification of blood regions for hemostasis
  support in laparoscopic surgery,'' {\em Signal, Image and Video Processing},
  vol.~13, no.~2, pp.~405--412, 2019.

\bibitem{li2009computer}
B.~Li and M.~Q.-H. Meng, ``Computer-aided detection of bleeding regions for
  capsule endoscopy images,'' {\em IEEE Transactions on biomedical
  engineering}, vol.~56, no.~4, pp.~1032--1039, 2009.

\bibitem{pan2011bleeding}
G.~Pan, G.~Yan, X.~Qiu, and J.~Cui, ``Bleeding detection in wireless capsule
  endoscopy based on probabilistic neural network,'' {\em Journal of medical
  systems}, vol.~35, no.~6, pp.~1477--1484, 2011.

\bibitem{fu2011bleeding}
Y.~Fu, M.~Mandal, and G.~Guo, ``Bleeding region detection in wce images based
  on color features and neural network,'' in {\em 2011 IEEE 54th International
  Midwest Symposium on Circuits and Systems (MWSCAS)}, pp.~1--4, IEEE, 2011.

\bibitem{mottaghi2017see}
R.~Mottaghi, C.~Schenck, D.~Fox, and A.~Farhadi, ``See the glass half full:
  Reasoning about liquid containers, their volume and content,'' in {\em
  Proceedings of the IEEE International Conference on Computer Vision},
  pp.~1871--1880, 2017.

\bibitem{schenck2018perceiving}
C.~Schenck and D.~Fox, ``Perceiving and reasoning about liquids using fully
  convolutional networks,'' {\em The International Journal of Robotics
  Research}, vol.~37, no.~4-5, pp.~452--471, 2018.

\bibitem{schenck2018spnets}
C.~Schenck and D.~Fox, ``Spnets: Differentiable fluid dynamics for deep neural
  networks,'' {\em arXiv preprint arXiv:1806.06094}, 2018.

\bibitem{yamaguchi2016stereo}
A.~Yamaguchi and C.~G. Atkeson, ``Stereo vision of liquid and particle flow for
  robot pouring,'' in {\em 2016 IEEE-RAS 16th International Conference on
  Humanoid Robots (Humanoids)}, pp.~1173--1180, IEEE, 2016.

\bibitem{allan20192017}
M.~Allan, A.~Shvets, T.~Kurmann, Z.~Zhang, R.~Duggal, Y.-H. Su, N.~Rieke,
  I.~Laina, N.~Kalavakonda, S.~Bodenstedt, {\em et~al.}, ``2017 robotic
  instrument segmentation challenge,'' {\em arXiv preprint arXiv:1902.06426},
  2019.

\bibitem{teney2016learning}
D.~Teney and M.~Hebert, ``Learning to extract motion from videos in
  convolutional neural networks,'' in {\em Asian Conference on Computer
  Vision}, pp.~412--428, Springer, 2016.

\bibitem{lucas1981iterative}
B.~D. Lucas, T.~Kanade, {\em et~al.}, ``An iterative image registration
  technique with an application to stereo vision,'' 1981.

\bibitem{baker2011database}
S.~Baker, D.~Scharstein, J.~Lewis, S.~Roth, M.~J. Black, and R.~Szeliski, ``A
  database and evaluation methodology for optical flow,'' {\em International
  journal of computer vision}, vol.~92, no.~1, pp.~1--31, 2011.

\bibitem{farneback2003two}
G.~Farneb{\"a}ck, ``Two-frame motion estimation based on polynomial
  expansion,'' in {\em Scandinavian conference on Image analysis},
  pp.~363--370, Springer, 2003.

\bibitem{chang2018pyramid}
J.-R. Chang and Y.-S. Chen, ``Pyramid stereo matching network,'' in {\em
  Proceedings of the IEEE Conference on Computer Vision and Pattern
  Recognition}, pp.~5410--5418, 2018.

\bibitem{ros}
{Stanford Artificial Intelligence Laboratory et al.}, ``Robotic operating
  system.''

\end{thebibliography}

\end{document}